\documentclass[conference]{IEEEtran}
\IEEEoverridecommandlockouts
\usepackage{cite}
\usepackage{amsmath,amssymb,amsfonts}
\usepackage{algorithmic}
\usepackage{graphicx}
\usepackage{textcomp}
\usepackage{xcolor}
\usepackage{booktabs}
\usepackage{placeins}
\usepackage{orcidlink}
\hypersetup{hidelinks}
\def\BibTeX{{\rm B\kern-.05em{\sc i\kern-.025em b}\kern-.08em
    T\kern-.1667em\lower.7ex\hbox{E}\kern-.125emX}}
\begin{document}
\bstctlcite{IEEEexample:BSTcontrol}

\title{When Morphs Look Better: Face Image Quality as a Training-Free Signal for Morphing Attack Detection
\thanks{Hao Peng, Fu Bo, and Mahmoud Elshirbeny contributed equally to this work.}
}

\author{\IEEEauthorblockN{Hao Peng\textsuperscript{1}\,\orcidlink{0009-0008-7188-1921},
Fu Bo\textsuperscript{1}\,\orcidlink{0009-0007-2500-7857},
Mahmoud Elshirbeny\textsuperscript{1,2}\,\orcidlink{0009-0007-3194-6869},
Guray Ozgur\textsuperscript{1,2}\,\orcidlink{0000-0002-1966-6641}}
\IEEEauthorblockA{\textsuperscript{1}TU Darmstadt, Darmstadt, Germany}
\IEEEauthorblockA{\textsuperscript{2}Fraunhofer Institute for Computer Graphics Research IGD, Darmstadt, Germany\\
\{hao.peng, fu.bo, mahmoud.elshirbeny\}@stud.tu-darmstadt.de, guray.ozgur@igd.fraunhofer.de}
}

\maketitle

\begin{abstract}
Face image quality assessment (FIQA) estimates the utility of a face image for automated face recognition and has been proposed as a morphing attack detection (MAD) signal that requires no morph-specific training. We evaluate 17 FIQA methods from all current FIQA categories on 20 morphing attack sets from two datasets, FRLL and FEI, including five diffusion-based attack sets. Each quality score is used directly as a MAD decision score. Unlike prior work, which inverted a quality measure when this lowered its error, we fix the score direction of each method in advance from its definition, so that cases in which morphs receive higher quality than bona fide samples remain visible. On FRLL, DifFIQA(R) achieves a mean D-EER of 11.42\%, and together with CLIB-FIQA and eDifFIQA(L) clearly outperforms all other FIQA methods. A strong separation of bona fide and morph scores does not imply good detection performance: on the diffusion-based attacks, up to 11 of 17 methods assign higher quality to morphs than to bona fide samples, and the largest reversed separation in the study coincides with the worst detection result of the study. Performance also depends strongly on the attack: GraFIQs(L), the best detector of DCMorph and Arc2Morph on FRLL, performs poorly on the landmark-based and GAN-based attacks. On FEI, the lowest mean D-EER is 39.75\%, and a threshold set on the bona fide data of one dataset does not transfer to the other.
\end{abstract}


\section{Introduction}
Face recognition (FR) is widely used for identity verification in travel documents and border control. A face morphing attack combines the face images of two or more subjects into one image that can be verified against each of them \cite{DBLP:conf/icb/FerraraFM14}, so that a document issued to one person can be used by another. Morphs can also be difficult for human examiners to detect \cite{DBLP:journals/tbbis/ZhangVRRDB21,DBLP:journals/corr/abs-2604-21627}, which makes morphing attack detection (MAD) a necessary part of secure FR systems. Single-image MAD is mostly addressed as supervised binary classification, from handcrafted descriptors to deep representations \cite{DBLP:journals/corr/abs-2011-02045}. The performance of such detectors depends on their training data and degrades when the morphing technique, the image source, or the post-processing differs at deployment \cite{DBLP:conf/iwbf/ScherhagRB18}. As morph generation has moved from landmark-based blending through GAN-based interpolation to identity-conditioned diffusion \cite{DBLP:journals/corr/abs-2604-21627}, detectors trained on known attacks are regularly confronted with new attack types. This motivates detection signals that are not learned from morphs. Face image quality assessment (FIQA) estimates the utility of a face image for recognition \cite{ISOIEC29794-1} and is trained without morphs. Since morphing alters texture, geometry, and identity information, it can shift the FIQA score. In that case, the score can be thresholded directly as a MAD score, without morph data or training, as only the threshold is set on bona fide images. Fu and Damer studied this with five FIQA and five general image quality measures on five morphing datasets \cite{DBLP:journals/iet-bmt/FuD22}. Since then, newer FIQA methods have been proposed in all categories, including methods based on the robustness to diffusion-based perturbation and methods that probe a pretrained FR model through gradients, intermediate layers, attention, or pruning, and diffusion-based morphing synthesizes the morph instead of blending two photographs. To our knowledge, quality-based MAD has not been evaluated with these FIQA methods or on diffusion-based attacks. We evaluate 17 FIQA methods from all categories, without retraining, on 20 morphing attack sets from two bona fide datasets, covering landmark-based, GAN-based, and diffusion-based morphing. Each FIQA score has a defined direction, and prior work inverted a quality measure when this lowered its average error over the attacks \cite{DBLP:journals/iet-bmt/FuD22}. This choice can present a method that assigns higher quality to morphs than to bona fide samples as a successful detector, and it requires knowledge of the attack. We instead fix the score direction of each method in advance and never change it per attack. Our contributions are threefold. (i) We evaluate 17 FIQA methods from all categories as training-free single-image MAD scores under one protocol and operating point, including five diffusion-based attack sets. (ii) We use a protocol with a fixed score direction and a signed separation measure, which shows that a strong separation of bona fide and morph scores does not imply good detection performance, most clearly on diffusion-based attacks. (iii) We analyze the limits of the approach: the FIQA performance of a method does not reliably predict its MAD performance, detection performance differs strongly between datasets, and a threshold set on bona fide data does not transfer between them.
\section{Related Work}
\label{sec:related}
\subsection{Face Image Quality Assessment}
\label{sec:related_fiqa}
FIQA estimates the utility of a face image for recognition rather than its perceived visual quality \cite{ISOIEC29794-1}. Following \cite{DBLP:journals/corr/abs-2605-13396}, we group FIQA methods into four categories. \emph{Supervised} methods regress a quality label: \textbf{FaceQnet} uses labels derived from ICAO compliance \cite{DBLP:conf/icb/Hernandez-Ortega19}, \textbf{SDD-FIQA} the distance between genuine and impostor similarity distributions \cite{DBLP:conf/cvpr/OuCZHLLLCW21}, and \textbf{CLIB-FIQA} quality anchors calibrated against facial attributes \cite{DBLP:conf/cvpr/OuL0K24}. \emph{Unsupervised} methods require no quality labels and form two categories. In \emph{non-FR-model} methods, quality comes from perturbing the input image with an external generative model rather than from the internals of the FR model: \textbf{DifFIQA} measures the resulting change of FR embeddings and distills it into a regression model, DifFIQA(R) \cite{DBLP:conf/icb/BabnikPS23}, and \textbf{eDifFIQA} improves this distillation \cite{DBLP:journals/tbbis/BabnikPS24}. \emph{FR-probing} methods derive quality from the internals of a frozen FR model, using the stability of embeddings under dropout (\textbf{SER-FIQ} \cite{DBLP:conf/cvpr/TerhorstKDKK20}), gradient magnitudes (\textbf{GraFIQs} \cite{DBLP:conf/cvpr/KolfDB22}), the stability of embeddings across ViT blocks (\textbf{ViTNT-FIQA} \cite{DBLP:conf/wacv/OzgurCCKHDB26}), informative intermediate layers (\textbf{FROQ} \cite{DBLP:conf/icb/BabnikJPS25}), the change of embeddings under pruning (\textbf{PreFIQs} \cite{DBLP:journals/corr/abs-2605-13396}), or pre-softmax self-attention (\textbf{ATTN-FIQA} \cite{DBLP:conf/fgr/OzgurCCKHADB26}). \emph{Self-supervised} methods derive quality from the FR training objective, either jointly with recognition or with an additional head on the FR model: \textbf{MagFace} through the embedding magnitude \cite{DBLP:conf/cvpr/MengZH021}, \textbf{PFE} through an uncertainty head on a frozen FR model \cite{DBLP:conf/iccv/ShiJ19}, \textbf{ViT-FIQA} through a learnable quality token \cite{DBLP:conf/iccv/AtzoriBD25}, \textbf{CR-FIQA} through relative classifiability \cite{DBLP:conf/cvpr/BoutrosFKFD23}, \textbf{CARPM-FIQA} through relative point margins accumulated over training \cite{ozgur2026carpm}, and \textbf{EX-FIQA} through early exits of ViT-FIQA \cite{DBLP:conf/fgr/OzgurCCKABD26}. These methods were developed and evaluated on bona fide face images, and none was designed for morphs, so whether and in which direction a method responds to morphing has to be measured. We include methods from all four categories and examine whether FIQA performance predicts MAD performance.
\subsection{Morphing Attacks and Their Detection}
\label{sec:related_mad}
Morph generation operates either on the image level or on a learned representation \cite{DBLP:journals/corr/abs-2604-21627}. We cover both with 20 attack sets: on FRLL, the five techniques of MAD22 \cite{DBLP:conf/icb/HuberBLRRDNGSCT22} and three diffusion-based techniques \cite{DBLP:journals/corr/abs-2604-21627,DBLP:conf/iwbf/DamerFSKHB23,DBLP:conf/fgr/DomenicoFFM26}, and the twelve techniques of FEI-Morph V3 \cite{DBLP:conf/iciap/DomenicoBFM23}. \emph{Image-level} methods detect facial landmarks in both images, interpolate them, and blend the texture: \textbf{OpenCV} uses 68 landmarks, Delaunay triangulation, and alpha blending \cite{Satya}. \textbf{FaceMorpher} is also landmark-based, but the version used in MAD22 is a commercial off-the-shelf tool whose technical details are not released \cite{DBLP:conf/icb/HuberBLRRDNGSCT22}. \textbf{WebMorph} uses 189 landmarks annotated for the underlying dataset, which reduces artifacts \cite{DeBruine}. Such morphs typically contain blending artifacts and geometric inconsistencies \cite{DBLP:journals/tbbis/ZhangVRRDB21}. \emph{Representation-level} methods interpolate in a learned latent space: \textbf{MIPGAN-I} and \textbf{MIPGAN-II} optimize perceptual quality, identity, identity difference, and structural objectives with StyleGAN and StyleGAN2, respectively, and produce high-resolution morphs with few visible artifacts \cite{DBLP:journals/tbbis/ZhangVRRDB21}. \textbf{DCMorph} is diffusion-based and injects identity features of both subjects into the denoising process via decoupled cross-attention, while DDIM inversion with spherical interpolation provides the initial latent \cite{DBLP:journals/corr/abs-2604-21627}. \textbf{MorDIFF} interpolates the semantic latent of a diffusion autoencoder \cite{DBLP:conf/iwbf/DamerFSKHB23}, and \textbf{Arc2Morph} interpolates the identity embeddings that condition the face generator Arc2Face \cite{DBLP:conf/fgr/DomenicoFFM26}. \textbf{LADIMO} \cite{DBLP:conf/icb/Grimmer024} and \textbf{StableMorph} \cite{DBLP:conf/icb/KabbaniRRB25} use latent diffusion and Stable Diffusion, respectively, and landmark-based morphing has been further improved as well \cite{DBLP:conf/iwbf/BatskosS24}. Unlike landmark-based morphs, diffusion-based morphs are synthesized from an interpolated conditioning signal rather than blended from two photographs. MAD is performed either differentially, using a trusted live capture, or on the suspected image alone, which is our setting. Both are surveyed in \cite{DBLP:journals/corr/abs-2011-02045}. Single-image MAD is mostly addressed as supervised binary classification, from handcrafted texture and forensic descriptors to deep networks trained with morph labels \cite{DBLP:journals/corr/abs-2011-02045}. Its main challenge is generalization, since detection cues depend on the morphing algorithm, the source data, compression, and re-digitization, and degrade when these are not seen in training \cite{DBLP:conf/iwbf/ScherhagRB18}. Proposed solutions extend the training data, treat morphs as anomalies without using morph labels \cite{DBLP:conf/icb/FangBD22}, or adapt foundation models, such as MADation \cite{DBLP:conf/wacv/CaldeiraOCIPBSD25} and the self-supervised SelfMAD++ \cite{DBLP:journals/inffus/IvanovskaTPS26}, but all of them train a model. In quality-based MAD, no model is trained, and the score of a pretrained FIQA method is used directly. Fu et al.\ showed that morphing shifts the quality distributions of morph images away from those of bona fide images \cite{DBLP:conf/biosig/FuSCD21}, and Fu and Damer evaluated this for MAD on five morphing datasets with ten quality measures \cite{DBLP:journals/iet-bmt/FuD22}. In differential MAD, quality has been used to support a separately trained detector rather than as the decision score \cite{DBLP:conf/icb/FrancoFL0M24,DBLP:conf/iwbf/SchlettRTB25}. SYN-MAD 2022 released the MAD22 benchmark and ranked submissions by the attack error at a fixed bona fide error of $20\%$ \cite{DBLP:conf/icb/HuberBLRRDNGSCT22}, i.e.\ MACER at BSCER $=20\%$ in the terms of ISO/IEC 20059 \cite{iso200592025}. Since all attacks share one bona fide set, fixing BSCER yields a single threshold and makes the attacks comparable. We adopt this benchmark and operating point and replace the trained detector with a pretrained FIQA score. Our work differs from prior quality-based MAD \cite{DBLP:journals/iet-bmt/FuD22} in two ways: it evaluated five FIQA methods, none of them from the non-FR-model category or among the recent FR-probing methods, whereas we evaluate 17 methods from all four categories, and it inverted a quality measure when this lowered its average EER over the attacks, whereas we fix the score direction in advance. In addition, quality-based MAD has, to our knowledge, only been evaluated on landmark-based and GAN-based attacks \cite{DBLP:conf/biosig/FuSCD21,DBLP:journals/iet-bmt/FuD22}, while we include five diffusion-based attack sets.
\section{Methodology}
\label{sec:methodology}
Each FIQA method is a pretrained function that maps an image to a scalar score, and we evaluate whether this score alone separates bona fide from morph images. Let $s_m(x)$ be the output of method $m$ for image $x$. The detector classifies $x$ as an attack if $q_m(x)\le\tau_m$, where $q_m$ is the direction-normalized score defined below. We consider the approach training-free if (i) the FIQA model is a released model trained for recognition utility and never exposed to morphs, (ii) no MAD classifier is trained and no method selection uses attack data, and (iii) no detector parameter, neither the score direction $\pi_m$ nor the threshold $\tau_m$, is derived from attacks. Attack labels are used only to measure performance. Condition (iii) distinguishes this protocol from prior work, which inverted a quality measure when this lowered its average EER over the attacks \cite{DBLP:journals/iet-bmt/FuD22}. FIQA methods differ in score direction: most scores increase with utility, while some increase with degradation, such as embedding uncertainty or gradient magnitude. Before any MAD evaluation, each method is therefore assigned a fixed $\pi_m\in\{+1,-1\}$ according to the meaning of its score in the original publication, and all analyses use $q_m(x)=\pi_m s_m(x)$, so that a larger $q_m$ always means higher predicted utility. $\pi_m$ is set per method and never changed per attack. Selecting the direction that minimizes the error, per method \cite{DBLP:journals/iet-bmt/FuD22} or per method and attack, lowers the reported error but requires knowledge of the attack and violates (iii). With a fixed direction, an error above $50\%$ occurs when morphing shifts the score opposite to its meaning. We first measure the effect of morphing independently of a threshold with the Fisher discriminant ratio (FDR) \cite{DBLP:journals/iet-bmt/FuD22} between the bona fide and attack score distributions of $q_m$,
\begin{equation}
\mathrm{FDR}=\frac{(\mu_{\mathrm{BF}}-\mu_{\mathrm{A}})^2}{\sigma_{\mathrm{BF}}^2+\sigma_{\mathrm{A}}^2},\qquad
S=\operatorname{sign}\!\left(\mu_{\mathrm{BF}}-\mu_{\mathrm{A}}\right)\sqrt{\mathrm{FDR}}.
\label{eq:fdr}
\end{equation}
As the mean difference is squared, FDR does not indicate the direction of the shift: morphs with higher quality than bona fide samples yield the same value as morphs with equally lower quality. We therefore report the signed separation $S$ in Tabs.~\ref{tab:mad_results} and~\ref{tab:fei}, whose magnitude is the separation and whose sign gives the direction: $S>0$ is the expected case, in which morphs receive lower quality, and $S<0$ a reversal, in which morphs receive higher quality. A large $|S|$ shows that a score responds to morphing, but only a large positive $S$ is useful for detection. The sign of $S$ depends on the means, whereas the D-EER depends on the ranking of the scores. In our results, they disagree, i.e.\ $S<0$ with a D-EER below $50\%$ or vice versa, only for $|S|<0.1$, where there is practically no separation. We use the metrics of ISO/IEC 20059 \cite{iso200592025}: the morphing attack classification error rate (MACER) is the proportion of morphs classified as bona fide, and the bona fide sample classification error rate (BSCER) is the proportion of bona fide samples classified as morphs. As threshold-independent measure, we report the detection equal error rate (D-EER), read at the ROC point closest to MACER $=$ BSCER, between the bona fide set and each attack set. Values near $50\%$ indicate no discrimination and values above $50\%$ a reversed ordering. Since the D-EER threshold is determined from both distributions, we also use a threshold that can be set in practice. $\tau_m$ is the lowest bona fide score at which BSCER reaches $20\%$ ($41/204=20.10\%$ on FRLL, $40/200=20\%$ on FEI). It is applied unchanged to all attacks of the same dataset, and we report the resulting MACER. This is the operating point of SYN-MAD 2022 \cite{DBLP:conf/icb/HuberBLRRDNGSCT22}, where the threshold is set on the bona fide images of the evaluated dataset. As BSCER is measured on the same bona fide images on which $\tau_m$ is set, BSCER $=20\%$ holds by construction, and MACER is an optimistic, in-sample estimate. Sec.~\ref{sec:transfer} therefore also tests the threshold on the other dataset. As BSCER is then identical for all attacks, MACER@BSCER$=20\%$ captures the attack-dependent error, and we report it instead of an average of both error rates. Since $\tau_m$ is computed from bona fide data only, it does not depend on the attacks and satisfies (iii). All metrics are computed per attack, and the mean D-EER and mean MACER over the attacks of a dataset summarize each method. We report $95\%$ confidence intervals of the mean D-EER from 1,000 bootstrap replicates. Each replicate draws identities with replacement, weights each bona fide image by the number of draws of its identity and each morph by the product of the draws of its two identities, so that morphs with an undrawn identity are excluded, and recomputes the D-EER of every attack and their mean. To test whether a threshold set on bona fide data transfers, we also apply the threshold of one bona fide dataset unchanged to the other.

\begin{table}[t]
\centering
\footnotesize
\setlength{\tabcolsep}{3pt}
\caption{Evaluated FIQA methods by category. $\pi_m$ is the fixed score direction of Sec.~\ref{sec:methodology}. $\pi_m=-1$ marks a score that falls as predicted recognition utility rises.}
\label{tab:methods}
\renewcommand{\arraystretch}{0.92}%
\resizebox{\linewidth}{!}{%
\begin{tabular}{@{}llc@{}}
\toprule
\textbf{Method} & \textbf{Quality signal} & $\boldsymbol{\pi_m}$ \\
\midrule
\multicolumn{3}{@{}l}{\textit{Supervised}} \\
FaceQnet \cite{DBLP:conf/icb/Hernandez-Ortega19} & ICAO compliance & $+1$ \\
SDD-FIQA \cite{DBLP:conf/cvpr/OuCZHLLLCW21} & similarity-distr.\ distance & $+1$ \\
CLIB-FIQA \cite{DBLP:conf/cvpr/OuL0K24} & calibrated anchors & $+1$ \\
\midrule
\multicolumn{3}{@{}l}{\textit{Unsupervised, non-FR-model}} \\
DifFIQA(R) \cite{DBLP:conf/icb/BabnikPS23} & diffusion robustness, regression & $+1$ \\
eDifFIQA(L) \cite{DBLP:journals/tbbis/BabnikPS24} & diffusion robustness, distillation & $+1$ \\
\midrule
\multicolumn{3}{@{}l}{\textit{Unsupervised, FR-probing}} \\
SER-FIQ \cite{DBLP:conf/cvpr/TerhorstKDKK20} & dropout stability & $+1$ \\
GraFIQs(L) \cite{DBLP:conf/cvpr/KolfDB22} & gradient magnitude & $-1$ \\
FROQ \cite{DBLP:conf/icb/BabnikJPS25} & intermediate layers & $+1$ \\
ViTNT-FIQA \cite{DBLP:conf/wacv/OzgurCCKHDB26} & ViT patch stability & $+1$ \\
PreFIQs \cite{DBLP:journals/corr/abs-2605-13396} & pruning drift & $-1$ \\
ATTN-FIQA \cite{DBLP:conf/fgr/OzgurCCKHADB26} & pre-softmax attention & $+1$ \\
\midrule
\multicolumn{3}{@{}l}{\textit{Self-supervised (FR-integrated)}} \\
MagFace \cite{DBLP:conf/cvpr/MengZH021} & embedding magnitude & $+1$ \\
PFE \cite{DBLP:conf/iccv/ShiJ19} & embedding uncertainty & $-1$ \\
CR-FIQA(L) \cite{DBLP:conf/cvpr/BoutrosFKFD23} & relative classifiability & $+1$ \\
ViT-FIQA(T) \cite{DBLP:conf/iccv/AtzoriBD25} & quality token & $+1$ \\
CARPM-FIQA(L) \cite{ozgur2026carpm} & accumulated point margin & $+1$ \\
EX-FIQA-FW(T) \cite{DBLP:conf/fgr/OzgurCCKABD26} & early-exit fusion & $+1$ \\
\bottomrule
\end{tabular}}
\end{table}

\begin{table}[t]
\centering
\footnotesize
\setlength{\tabcolsep}{3pt}
\caption{Evaluated protocol. Every attack set is compared against the bona fide images of its own dataset.}
\label{tab:protocol}
\resizebox{\linewidth}{!}{%
\begin{tabular}{@{}llr@{}}
\toprule
\textbf{Subset} & \textbf{Family} & \textbf{Img.} \\
\midrule
FRLL bona fide \cite{DBLP:conf/icb/HuberBLRRDNGSCT22} & - & 204 \\
\midrule
FaceMorpher \cite{DBLP:conf/icb/HuberBLRRDNGSCT22} & landmark & 1,000 \\
OpenCV \cite{Satya} & landmark & 984 \\
WebMorph \cite{DeBruine} & landmark & 500 \\
MIPGAN-I \cite{DBLP:journals/tbbis/ZhangVRRDB21} & GAN & 1,000 \\
MIPGAN-II \cite{DBLP:journals/tbbis/ZhangVRRDB21} & GAN & 999 \\
DCMorph \cite{DBLP:journals/corr/abs-2604-21627} & diffusion & 1,000 \\
MorDIFF \cite{DBLP:conf/iwbf/DamerFSKHB23} & diffusion & 1,000 \\
Arc2Morph \cite{DBLP:conf/fgr/DomenicoFFM26} & diffusion & 1,000 \\
\midrule
FEI bona fide \cite{DBLP:journals/ivc/ThomazG10} & - & 200 \\
FEI-Morph V3, 8 sets & landmark / commercial & 8,000 \\
FEI-Morph V3, 4 sets & generative & 3,999 \\
\midrule
\textbf{Total attacks} & & \textbf{19,482} \\
\bottomrule
\end{tabular}}
\end{table}

\section{Experimental Details}
\label{sec:experimental}
We use two bona fide datasets (Tab.~\ref{tab:protocol}). The first is MAD22 \cite{DBLP:conf/icb/HuberBLRRDNGSCT22}, extended with three diffusion-based attacks on the same pairs: DCMorph, created by its authors \cite{DBLP:journals/corr/abs-2604-21627}, and MorDIFF \cite{DBLP:conf/iwbf/DamerFSKHB23} and Arc2Morph \cite{DBLP:conf/fgr/DomenicoFFM26}, which we generate with the authors' code and default settings. We refer to MAD22 extended with these three attacks as FRLL and to FEI-Morph V3 as FEI. MAD22 contains 204 frontal bona fide images of 102 subjects from the Face Research Lab London (FRLL) dataset and 1,000 morphing pairs, selected as the 250 most similar pairs by ElasticFace-Arc cosine similarity per gender and expression \cite{DBLP:conf/icb/HuberBLRRDNGSCT22}. The number of attacks differs per technique: 16 OpenCV morphs with mouth-region artifacts were removed, WebMorph was applied only to neutral images since it failed on smiling faces, and landmark detection failed on one MIPGAN-II morph \cite{DBLP:conf/icb/HuberBLRRDNGSCT22}. The second dataset is FEI-Morph V3 \cite{DBLP:conf/iciap/DomenicoBFM23}, based on the FEI face database \cite{DBLP:journals/ivc/ThomazG10}: 200 frontal bona fide images of 200 subjects and twelve morphing techniques, eight landmark-based or commercial \cite{Alyssa,DBLP:journals/access/BorghiDFFM23,DBLP:journals/fcomp/BatskosSV23,DBLP:conf/iwbf/BatskosS24} and four generative. Two of the generative ones are diffusion-based, namely Arc2Morph and StableMorph \cite{DBLP:conf/icb/KabbaniRRB25}, and two are described only as inspired by LADIMO \cite{DBLP:conf/icb/Grimmer024} and by the StyleGAN2-based demorphing method StyleDemorpher \cite{DBLP:journals/mva/IsmayilovSB25}. The providers do not document how these two sets produce morphs, and the LADIMO-inspired set contains 999 morphs. The FEI FaceMorpher and Arc2Morph sets may differ in implementation or settings from their FRLL counterparts, and SYN-MAD 2022 describes the MAD22 FaceMorpher as a commercial off-the-shelf tool. Within each dataset, all attack sets are evaluated against the same bona fide images, so that one threshold derived from bona fide data applies to all its attacks. Faces are detected and aligned to $112\times112$ pixels with MTCNN \cite{DBLP:journals/spl/ZhangZLQ16}. Further preprocessing follows each released implementation. No morph-specific preprocessing or attack-dependent tuning is applied. We evaluate the 17 methods in Tab.~\ref{tab:methods}, which cover all four categories of Sec.~\ref{sec:related_fiqa}. Each is a released model with its original weights and is neither retrained nor calibrated on morphing data. Tab.~\ref{tab:methods} also lists $\pi_m$: fourteen scores increase with predicted utility and are used unchanged, and three measure a degradation-related quantity and are inverted, namely PFE (embedding uncertainty \cite{DBLP:conf/iccv/ShiJ19}), GraFIQs (a gradient magnitude that increases as a sample deviates from the training distribution \cite{DBLP:conf/cvpr/KolfDB22}), and PreFIQs (the change of embeddings under pruning \cite{DBLP:journals/corr/abs-2605-13396}). SER-FIQ is stochastic, as its scores are computed from repeated dropout passes, so its results carry a sampling variance. MagFace and CR-FIQA(L) are min-max normalized with constants from an external dataset. This affine transform changes neither $S$ nor the error rates. The suffixes denote the evaluated variants: DifFIQA(R) is the regression model distilled from DifFIQA, eDifFIQA(L) the variant with the large backbone, GraFIQs(L), CR-FIQA(L), and CARPM-FIQA(L) use iResNet-100 backbones trained on MS1MV2, ViT-FIQA(T) is the quality-token variant, and EX-FIQA-FW(T) fuses all twelve early exits of ViT-FIQA(T) with depth-dependent weights. The FR-probing methods use the FR models released with them: an ArcFace ResNet-100 for SER-FIQ, GraFIQs(L), and PreFIQs, an AdaFace IR-101 for FROQ, and AdaFace ViT-B and ViT-S models trained on WebFace4M for ViTNT-FIQA and ATTN-FIQA, respectively. For reference only, Tabs.~\ref{tab:mad_results} and~\ref{tab:fei} also list the trained detector SelfMAD++ \cite{DBLP:journals/inffus/IvanovskaTPS26} with its released weights and preprocessing, applied to the same aligned images, which its face crop upsamples to its $512\times512$ input. As this is not its native input and may be suboptimal for it, its results only indicate the performance of a trained detector on the same images and are not compared with those of the FIQA methods.

\begin{table*}[t]
\centering
\caption{MAD performance on FRLL using FIQA scores directly as decision scores. Each attack cell reports the D-EER (\%) and, in parentheses, the signed separation $S$ of Eq.~(\ref{eq:fdr}). $S<0$ marks a reversal, i.e.\ morphs receive a higher mean quality than bona fide samples. A large $|S|$ indicates a strong separation, which is useful for detection only if $S>0$. Mean D-EER averages the eight D-EERs, and mean MACER (\%) is measured at the fixed bona fide operating point (BSCER $\approx 20\%$). Lower D-EER and MACER indicate better detection, and bold marks the best FIQA method per column. Values of $S$ are rounded, and its sign is that of the unrounded value. SelfMAD++ is a trained detector listed for reference only, without $S$, and is not compared (Sec.~\ref{sec:experimental}).}
\label{tab:mad_results}
\footnotesize
\setlength{\tabcolsep}{3pt}
\renewcommand{\arraystretch}{0.92}%
\resizebox{\textwidth}{!}{%
\begin{tabular}{@{}lcccccccccc@{}}
\toprule
& \multicolumn{3}{c}{\textbf{Landmark}} & \multicolumn{2}{c}{\textbf{GAN}} & \multicolumn{3}{c}{\textbf{Diffusion}} & & \\
\cmidrule(lr){2-4}\cmidrule(lr){5-6}\cmidrule(lr){7-9}
\textbf{Method} & \textbf{FM} & \textbf{OCV} & \textbf{WM} & \textbf{MI} & \textbf{MII} & \textbf{DCM} & \textbf{MD} & \textbf{A2M} & \textbf{Mean D-EER} & \textbf{Mean MACER} \\
\midrule
\multicolumn{11}{@{}l}{\textit{Reference only, not for comparison: trained detector on $112\times112$ inputs}} \\
SelfMAD++ \cite{DBLP:journals/inffus/IvanovskaTPS26} & 0.30 & 1.42 & 9.00 & 6.50 & 1.60 & 48.60 & 1.70 & 35.25 & 13.05 & 14.92 \\
\midrule
\multicolumn{11}{@{}l}{\textit{Supervised}} \\
FaceQnet & 46.50 \tiny(0.12) & 51.63 \tiny($-$0.05) & 48.00 \tiny(0.09) & 50.10 \tiny($-$0.03) & 49.65 \tiny($-$0.02) & 48.20 \tiny(0.05) & 51.40 \tiny($-$0.08) & 41.62 \tiny(0.25) & 48.39 & 81.25 \\
SDD-FIQA & 37.30 \tiny(0.45) & 49.29 \tiny($-$0.03) & 39.20 \tiny(0.36) & 33.00 \tiny(0.62) & 31.23 \tiny(0.70) & 66.60 \tiny($-$0.59) & 67.90 \tiny($-$0.63) & 51.72 \tiny($-$0.06) & 47.03 & 70.52 \\
CLIB-FIQA & 2.30 \tiny(2.82) & 24.80 \tiny(0.90) & 27.60 \tiny(0.76) & 11.70 \tiny(1.72) & \textbf{8.21} \tiny(1.91) & 48.00 \tiny(0.08) & 3.10 \tiny(2.50) & 10.20 \tiny(1.85) & 16.99 & 20.03 \\
\midrule
\multicolumn{11}{@{}l}{\textit{Unsupervised, non-FR-model}} \\
DifFIQA(R) & \textbf{0.00} \tiny(3.52) & \textbf{10.47} \tiny(1.76) & \textbf{21.00} \tiny(1.13) & 14.00 \tiny(1.56) & \textbf{8.21} \tiny(1.84) & 30.00 \tiny(0.74) & \textbf{1.30} \tiny(3.28) & 6.36 \tiny(2.12) & \textbf{11.42} & \textbf{10.75} \\
eDifFIQA(L) & 3.40 \tiny(2.25) & 24.09 \tiny(0.97) & 24.80 \tiny(0.89) & \textbf{8.30} \tiny(1.96) & 9.51 \tiny(1.89) & 63.20 \tiny($-$0.53) & 12.00 \tiny(1.63) & 49.39 \tiny(0.07) & 24.34 & 30.47 \\
\midrule
\multicolumn{11}{@{}l}{\textit{Unsupervised, FR-probing}} \\
SER-FIQ & 47.40 \tiny(0.12) & 50.91 \tiny($-$0.10) & 52.60 \tiny($-$0.14) & 36.90 \tiny(0.37) & 36.34 \tiny(0.36) & 57.00 \tiny($-$0.22) & 42.10 \tiny(0.16) & 62.22 \tiny($-$0.39) & 48.18 & 80.42 \\
GraFIQs(L) & 51.10 \tiny(0.06) & 49.80 \tiny($-$0.09) & 61.20 \tiny($-$0.50) & 37.70 \tiny(0.40) & 39.84 \tiny(0.35) & \textbf{9.20} \tiny(1.89) & 16.30 \tiny(1.36) & \textbf{5.66} \tiny(2.19) & 33.85 & 55.66 \\
FROQ & 31.60 \tiny(0.74) & 47.26 \tiny(0.12) & 47.20 \tiny(0.11) & 27.50 \tiny(0.82) & 25.93 \tiny(0.91) & 94.90 \tiny($-$2.29) & 83.00 \tiny($-$1.28) & 94.14 \tiny($-$2.14) & 56.44 & 73.19 \\
ViTNT-FIQA & 71.90 \tiny($-$0.51) & 71.04 \tiny($-$0.74) & 70.80 \tiny($-$0.60) & 64.30 \tiny($-$0.47) & 56.56 \tiny($-$0.28) & 36.70 \tiny(0.39) & 59.80 \tiny($-$0.33) & 21.31 \tiny(0.96) & 56.55 & 80.40 \\
PreFIQs & 44.70 \tiny(0.30) & 48.98 \tiny(0.08) & 51.80 \tiny(0.00) & 28.80 \tiny(0.75) & 28.63 \tiny(0.76) & 46.20 \tiny(0.17) & 41.10 \tiny(0.33) & 51.41 \tiny(0.00) & 42.70 & 61.84 \\
ATTN-FIQA & 40.60 \tiny(0.42) & 55.08 \tiny($-$0.10) & 52.80 \tiny($-$0.02) & 48.50 \tiny(0.11) & 39.64 \tiny(0.39) & 56.50 \tiny($-$0.15) & 60.40 \tiny($-$0.32) & 50.10 \tiny(0.06) & 50.45 & 79.31 \\
\midrule
\multicolumn{11}{@{}l}{\textit{Self-supervised (FR-integrated)}} \\
MagFace & 29.90 \tiny(0.85) & 41.16 \tiny(0.31) & 42.00 \tiny(0.42) & 20.00 \tiny(1.25) & 19.22 \tiny(1.26) & 59.00 \tiny($-$0.31) & 42.60 \tiny(0.27) & 81.11 \tiny($-$1.19) & 41.87 & 59.44 \\
PFE & 27.00 \tiny(0.87) & 40.65 \tiny(0.30) & 31.20 \tiny(0.71) & 26.80 \tiny(0.88) & 23.02 \tiny(1.05) & 75.30 \tiny($-$0.89) & 52.10 \tiny($-$0.05) & 60.20 \tiny($-$0.41) & 42.03 & 61.85 \\
CR-FIQA(L) & 33.50 \tiny(0.59) & 47.05 \tiny(0.17) & 36.20 \tiny(0.50) & 15.10 \tiny(1.27) & 15.12 \tiny(1.27) & 67.90 \tiny($-$0.57) & 59.70 \tiny($-$0.36) & 71.21 \tiny($-$0.83) & 43.22 & 58.03 \\
ViT-FIQA(T) & 29.00 \tiny(0.76) & 41.77 \tiny(0.28) & 30.00 \tiny(0.70) & 27.10 \tiny(0.89) & 23.62 \tiny(1.06) & 78.40 \tiny($-$0.97) & 61.00 \tiny($-$0.38) & 58.28 \tiny($-$0.26) & 43.65 & 65.65 \\
CARPM-FIQA(L) & 32.50 \tiny(0.64) & 43.70 \tiny(0.22) & 33.40 \tiny(0.61) & 16.30 \tiny(1.32) & 12.31 \tiny(1.48) & 67.50 \tiny($-$0.54) & 56.60 \tiny($-$0.23) & 66.36 \tiny($-$0.58) & 41.08 & 56.93 \\
EX-FIQA-FW(T) & 41.00 \tiny(0.39) & 54.17 \tiny($-$0.17) & 44.00 \tiny(0.17) & 42.10 \tiny(0.28) & 37.24 \tiny(0.45) & 69.90 \tiny($-$0.70) & 66.60 \tiny($-$0.61) & 40.91 \tiny(0.29) & 49.49 & 82.37 \\
\bottomrule
\end{tabular}}
{\footnotesize FM: FaceMorpher, OCV: OpenCV, WM: WebMorph, MI: MIPGAN-I, MII: MIPGAN-II, DCM: DCMorph, MD: MorDIFF, A2M: Arc2Morph.}
\end{table*}

\begin{table*}[t]
\centering
\caption{MAD performance on FEI-Morph V3: D-EER (\%) per attack with the signed separation $S$ in parentheses, mean D-EER, and mean MACER (\%) at BSCER $=20\%$. Methods, $S$, and bold as in Tab.~\ref{tab:mad_results}.}
\label{tab:fei}
\scriptsize
\setlength{\tabcolsep}{2.2pt}
\renewcommand{\arraystretch}{0.92}%
\resizebox{\textwidth}{!}{%
\begin{tabular}{@{}lcccccccccccccc@{}}
\toprule
& \multicolumn{8}{c}{\textbf{Landmark / commercial}} & \multicolumn{4}{c}{\textbf{Generative}} & & \\
\cmidrule(lr){2-9}\cmidrule(lr){10-13}
\textbf{Method} & \textbf{FM} & \textbf{FF} & \textbf{NTNU} & \textbf{UBO} & \textbf{STA} & \textbf{SUR} & \textbf{TwB} & \textbf{TwI} & \textbf{LAD} & \textbf{SDM} & \textbf{A2M} & \textbf{StM} & \textbf{Mean D-EER} & \textbf{Mean MACER} \\
\midrule
\multicolumn{15}{@{}l}{\textit{Reference only, not for comparison: trained detector on $112\times112$ inputs}} \\
SelfMAD++ \cite{DBLP:journals/inffus/IvanovskaTPS26} & 0.0 & 18.7 & 27.0 & 18.4 & 46.3 & 14.8 & 38.4 & 60.8 & 11.5 & 11.2 & 91.9 & 15.7 & 29.56 & 35.57 \\
\midrule
\multicolumn{15}{@{}l}{\textit{Supervised}} \\
FaceQnet & 49.6 \tiny($-$0.01) & 48.7 \tiny(0.01) & 47.6 \tiny(0.02) & 49.0 \tiny(0.03) & 49.0 \tiny(0.00) & 48.0 \tiny(0.04) & 49.3 \tiny($-$0.01) & 48.5 \tiny(0.03) & 39.3 \tiny(0.36) & 54.7 \tiny($-$0.15) & 44.8 \tiny(0.18) & 44.6 \tiny(0.19) & 47.76 & 78.54 \\
SDD-FIQA & 46.5 \tiny(0.13) & 49.1 \tiny(0.07) & 47.2 \tiny(0.08) & 48.6 \tiny(0.05) & 48.0 \tiny(0.09) & 49.1 \tiny(0.07) & 45.4 \tiny(0.15) & 44.0 \tiny(0.20) & 35.1 \tiny(0.56) & 52.0 \tiny($-$0.08) & 49.4 \tiny(0.03) & 49.9 \tiny(0.01) & 47.02 & 78.53 \\
CLIB-FIQA & 28.1 \tiny(0.79) & 35.3 \tiny(0.48) & 33.2 \tiny(0.56) & 35.5 \tiny(0.49) & 44.9 \tiny(0.11) & 35.8 \tiny(0.44) & 42.7 \tiny(0.21) & 49.4 \tiny($-$0.04) & 46.5 \tiny(0.06) & 42.0 \tiny(0.25) & 87.5 \tiny($-$1.60) & 57.0 \tiny($-$0.30) & 44.82 & 71.19 \\
\midrule
\multicolumn{15}{@{}l}{\textit{Unsupervised, non-FR-model}} \\
DifFIQA(R) & \textbf{22.6} \tiny(1.03) & \textbf{31.3} \tiny(0.62) & \textbf{31.1} \tiny(0.60) & 31.3 \tiny(0.61) & 43.3 \tiny(0.23) & 35.2 \tiny(0.48) & 41.6 \tiny(0.27) & 56.0 \tiny($-$0.22) & 51.5 \tiny($-$0.10) & 27.8 \tiny(0.79) & 88.4 \tiny($-$1.56) & 70.9 \tiny($-$0.83) & 44.25 & 66.03 \\
eDifFIQA(L) & 24.9 \tiny(0.98) & 32.3 \tiny(0.65) & 32.2 \tiny(0.66) & \textbf{31.2} \tiny(0.72) & 40.4 \tiny(0.33) & \textbf{31.4} \tiny(0.68) & 39.3 \tiny(0.38) & 45.0 \tiny(0.18) & 40.9 \tiny(0.35) & \textbf{24.1} \tiny(1.01) & 79.2 \tiny($-$1.29) & 69.2 \tiny($-$0.71) & 40.84 & 60.28 \\
\midrule
\multicolumn{15}{@{}l}{\textit{Unsupervised, FR-probing}} \\
SER-FIQ & 48.6 \tiny(0.03) & 53.4 \tiny($-$0.12) & 54.5 \tiny($-$0.16) & 54.6 \tiny($-$0.12) & 53.0 \tiny($-$0.09) & 54.8 \tiny($-$0.17) & 52.4 \tiny($-$0.06) & 50.9 \tiny($-$0.02) & 62.1 \tiny($-$0.46) & 41.9 \tiny(0.29) & 52.8 \tiny($-$0.06) & 53.9 \tiny($-$0.07) & 52.74 & 83.39 \\
GraFIQs(L) & 49.4 \tiny(0.02) & 53.0 \tiny($-$0.10) & 46.8 \tiny(0.08) & 47.3 \tiny(0.08) & 54.3 \tiny($-$0.16) & 46.0 \tiny(0.11) & 50.3 \tiny($-$0.05) & 56.9 \tiny($-$0.28) & 58.6 \tiny($-$0.28) & 43.2 \tiny(0.20) & 22.9 \tiny(1.06) & 22.9 \tiny(0.89) & 45.97 & 74.38 \\
FROQ & 35.2 \tiny(0.42) & 45.0 \tiny(0.13) & 49.5 \tiny($-$0.02) & 49.6 \tiny(0.01) & 44.2 \tiny(0.15) & 47.8 \tiny(0.05) & 45.0 \tiny(0.14) & 50.9 \tiny(0.00) & 39.7 \tiny(0.29) & 66.1 \tiny($-$0.62) & 81.0 \tiny($-$1.25) & 51.5 \tiny($-$0.05) & 50.46 & 80.03 \\
ViTNT-FIQA & 66.0 \tiny($-$0.60) & 63.8 \tiny($-$0.55) & 61.8 \tiny($-$0.43) & 63.4 \tiny($-$0.51) & 59.9 \tiny($-$0.34) & 58.2 \tiny($-$0.30) & 56.8 \tiny($-$0.23) & 50.9 \tiny($-$0.08) & 47.1 \tiny(0.13) & 71.9 \tiny($-$0.83) & \textbf{10.0} \tiny(1.81) & \textbf{20.9} \tiny(1.08) & 52.56 & 76.91 \\
PreFIQs & 47.3 \tiny(0.13) & 50.1 \tiny(0.05) & 49.8 \tiny(0.05) & 49.2 \tiny(0.08) & 47.7 \tiny(0.07) & 48.1 \tiny(0.11) & 48.1 \tiny(0.10) & 47.1 \tiny(0.12) & 57.2 \tiny($-$0.23) & 36.9 \tiny(0.43) & 56.2 \tiny($-$0.16) & 55.0 \tiny($-$0.11) & 49.39 & 78.55 \\
ATTN-FIQA & 52.8 \tiny($-$0.06) & 56.5 \tiny($-$0.20) & 54.0 \tiny($-$0.13) & 57.0 \tiny($-$0.24) & 49.0 \tiny(0.03) & 47.7 \tiny(0.07) & 48.4 \tiny(0.05) & 44.3 \tiny(0.12) & 31.1 \tiny(0.69) & 56.3 \tiny($-$0.19) & 39.9 \tiny(0.31) & 39.3 \tiny(0.32) & 48.02 & 79.41 \\
\midrule
\multicolumn{15}{@{}l}{\textit{Self-supervised (FR-integrated)}} \\
MagFace & 33.9 \tiny(0.60) & 40.6 \tiny(0.35) & 40.3 \tiny(0.37) & 40.7 \tiny(0.37) & 37.4 \tiny(0.47) & 36.1 \tiny(0.50) & \textbf{34.3} \tiny(0.55) & \textbf{33.7} \tiny(0.62) & 44.7 \tiny(0.17) & 30.1 \tiny(0.74) & 66.0 \tiny($-$0.66) & 48.8 \tiny(0.03) & 40.55 & 61.91 \\
PFE & 28.1 \tiny(0.78) & 38.4 \tiny(0.44) & 36.1 \tiny(0.52) & 38.5 \tiny(0.45) & 37.0 \tiny(0.48) & 37.5 \tiny(0.48) & 35.4 \tiny(0.53) & 38.4 \tiny(0.45) & 26.9 \tiny(0.83) & 43.6 \tiny(0.20) & 66.1 \tiny($-$0.48) & 71.4 \tiny($-$0.75) & 41.45 & 61.40 \\
CR-FIQA(L) & 41.1 \tiny(0.43) & 46.0 \tiny(0.20) & 47.0 \tiny(0.18) & 47.8 \tiny(0.14) & 43.8 \tiny(0.26) & 44.0 \tiny(0.29) & 45.1 \tiny(0.26) & 42.4 \tiny(0.34) & 33.6 \tiny(0.64) & 43.7 \tiny(0.30) & 63.0 \tiny($-$0.53) & 55.2 \tiny($-$0.07) & 46.06 & 71.23 \\
ViT-FIQA(T) & 30.5 \tiny(0.71) & 39.6 \tiny(0.37) & 39.2 \tiny(0.41) & 41.3 \tiny(0.31) & \textbf{36.0} \tiny(0.55) & 36.8 \tiny(0.46) & 38.6 \tiny(0.51) & 36.4 \tiny(0.49) & \textbf{20.9} \tiny(1.01) & 41.5 \tiny(0.30) & 48.7 \tiny(0.02) & 67.5 \tiny($-$0.62) & \textbf{39.75} & \textbf{58.96} \\
CARPM-FIQA(L) & 36.8 \tiny(0.50) & 44.5 \tiny(0.25) & 42.0 \tiny(0.28) & 44.8 \tiny(0.23) & 42.7 \tiny(0.30) & 41.8 \tiny(0.32) & 43.1 \tiny(0.29) & 41.1 \tiny(0.35) & 34.7 \tiny(0.53) & 38.9 \tiny(0.42) & 58.7 \tiny($-$0.32) & 50.8 \tiny($-$0.02) & 43.32 & 69.31 \\
EX-FIQA-FW(T) & 39.1 \tiny(0.34) & 44.8 \tiny(0.13) & 43.1 \tiny(0.20) & 46.0 \tiny(0.07) & 39.4 \tiny(0.36) & 40.8 \tiny(0.27) & 39.6 \tiny(0.36) & 41.7 \tiny(0.26) & 27.3 \tiny(0.77) & 53.5 \tiny($-$0.18) & 29.9 \tiny(0.72) & 44.8 \tiny(0.13) & 40.83 & 70.56 \\
\bottomrule
\end{tabular}}
{\footnotesize FM: FaceMorpher, FF: FaceFusion, NTNU: NTNU FaceMorph, UBO: UBO Morph, STA: Twente STASM variant, SUR: SURYS, TwB: Twente \cite{DBLP:journals/fcomp/BatskosSV23}, TwI: improved Twente landmark morphing \cite{DBLP:conf/iwbf/BatskosS24}, LAD/SDM: LADIMO-/StyleDemorpher-inspired, A2M: Arc2Morph, StM: StableMorph.}
\end{table*}

\section{Results}
\label{sec:results}
Tab.~\ref{tab:mad_results} reports, for each FRLL attack, the signed separation $S$ of bona fide and morph scores. Its magnitude ranges from $0.00$ to $3.52$ across methods and attacks. DifFIQA(R) separates FaceMorpher with $S=3.52$ and CLIB-FIQA with $S=2.82$, while $|S|$ does not exceed $0.25$ for FaceQnet on any FRLL attack. Separation varies strongly within a category: among the FR-probing methods, FROQ reaches $|S|=2.29$ on DCMorph, while $|S|$ does not exceed $0.39$ for SER-FIQ on any FRLL attack. Separation also depends on the attack: OpenCV and WebMorph are weakly separated by almost all methods, with $|S|>1$ only for DifFIQA(R), while FaceMorpher and both MIPGAN attacks are separated more clearly. The direction of the shift differs between methods, and reversals ($S<0$) occur for some methods on landmark-based and GAN-based attacks but are most frequent on diffusion-based attacks: 11, 10, and 8 of 17 methods reverse on DCMorph, MorDIFF, and Arc2Morph, and on FEI 10 of 17 reverse on Arc2Morph and StableMorph, compared with at most 5 on any landmark-based FEI set. The LADIMO- and StyleDemorpher-inspired sets show 4 and 6 reversals. One hypothesis is that morphs without blending artifacts receive a higher quality. The GAN-based MIPGAN attacks, which contain no blending artifacts either, show only 2 reversals each, so this hypothesis alone does not explain the reversals. On DCMorph, FROQ reaches the largest reversed separation of the study, $S=-2.29$ ($-2.14$ on Arc2Morph), while GraFIQs(L) reaches $S=+1.89$ in the expected direction.

\subsection{Detection Performance of FIQA Scores}
\label{sec:mad_results}
Tab.~\ref{tab:mad_results} also reports the D-EER per attack and, per method, the mean D-EER and the mean MACER at the fixed operating point. On FRLL, DifFIQA(R) achieves the best results with a mean D-EER of $11.42\%$ (confidence interval $[9.0, 14.0]$, below the mean of every other method) and a mean MACER of $10.75\%$, followed by CLIB-FIQA and eDifFIQA(L) with $16.99/20.03\%$ and $24.34/30.47\%$ (mean D-EER/mean MACER), while the fourth-ranked method by both mean D-EER and mean MACER, GraFIQs(L), reaches a mean MACER of $55.66\%$. Quality-based MAD is thus effective on this dataset without morph data or training, with the threshold set on its bona fide images. Identifying DifFIQA(R) as the best method, however, uses the attack labels of this evaluation, and how a deployer could select a method without attack data remains open. Performance depends strongly on the attack. All three reach their lowest D-EER on FaceMorpher ($0.00$-$3.40\%$) and their highest on DCMorph ($30.00$-$63.20\%$), and eDifFIQA(L) is also close to chance on Arc2Morph ($49.39\%$). PFE, ViT-FIQA(T), and FROQ reach DCMorph D-EERs of $75.30\%$, $78.40\%$, and $94.90\%$, well above $50\%$ and consistent with their negative $S$ in Tab.~\ref{tab:mad_results}. FROQ separates DCMorph more strongly than any other reversed method-attack pair but is the worst detector of this attack. If the direction were selected per attack, its DCMorph D-EER would be $5.10\%$, lower than the best DCMorph result in Tab.~\ref{tab:mad_results} ($9.20\%$, GraFIQs(L)). Conversely, GraFIQs(L) detects the landmark-based and GAN-based attacks poorly, with D-EERs of $37.70$-$61.20\%$, but reaches the lowest D-EERs on DCMorph ($9.20\%$) and Arc2Morph ($5.66\%$). On FEI (Tab.~\ref{tab:fei}), ViTNT-FIQA, last on FRLL and 16th of 17 on FEI, likewise detects Arc2Morph and StableMorph best ($10.0\%$ and $20.9\%$), while DifFIQA(R) reverses on both ($88.4\%$ and $70.9\%$). A method can thus perform poorly on most attacks and still be the best detector of others, which the mean over attacks does not show. Selecting the direction per attack would report these cases as successful detection but requires knowledge of the attack, contradicting the training-free setting. The direction is also not consistent for the same method and morphing technique across datasets: DifFIQA(R) detects FRLL Arc2Morph with $6.36\%$ but reverses on FEI Arc2Morph, although the two sets may differ in generation settings (Sec.~\ref{sec:experimental}). The reference detector SelfMAD++ also has its highest FRLL D-EERs on DCMorph ($48.60\%$) and Arc2Morph ($35.25\%$) and exceeds $50\%$ on FEI Arc2Morph ($91.9\%$). ViTNT-FIQA and FROQ, two of the most recent methods, rank last on FRLL, behind the oldest method, FaceQnet. PreFIQs, which reports state-of-the-art FIQA performance \cite{DBLP:journals/corr/abs-2605-13396}, ranks eighth of seventeen on FRLL and fourteenth on FEI. To compare FIQA and MAD performance systematically, we use our own FIQA evaluation of 14 methods (all except FROQ, PreFIQs, and EX-FIQA-FW(T)). It ranks the methods by the partial area under the error-versus-discard curve in each of the 64 combinations of four FR models, eight datasets, and two false match rates, and averages the ranks. This ranking correlates only moderately with the ranking of the same 14 methods by mean D-EER (Spearman $\rho=0.54$, $p=0.047$ on FRLL and $\rho=0.55$, $p=0.043$ on FEI). In a leave-one-out analysis, removing a single method makes the correlation non-significant for 8 (FRLL) and 9 (FEI) of the 14 methods, and individual methods differ by up to eight positions between the FIQA ranking and the ranking by mean D-EER. A higher FIQA ranking therefore does not reliably indicate better MAD performance.

\subsection{Transfer Across Datasets}
\label{sec:transfer}
Detection performance differs strongly between FRLL and FEI (Tab.~\ref{tab:fei}). Separation is weak, with $|S|$ not exceeding $1.03$ on any landmark-based FEI set. The lowest mean D-EER is $39.75\%$ and the lowest mean MACER $58.96\%$, both achieved by ViT-FIQA(T), whose D-EER confidence interval $[37.3, 42.6]$ includes the means of the next four methods (MagFace, EX-FIQA-FW(T), eDifFIQA(L), and PFE), while DifFIQA(R), the best method on FRLL, reaches a mean D-EER of $44.25\%$. As the attack sets differ, this does not isolate the effect of the bona fide data, which the following threshold experiment does. Set at BSCER $=20\%$ on FRLL, the thresholds of DifFIQA(R) and CLIB-FIQA yield a BSCER of $100\%$ on FEI. Set on FEI, they yield a BSCER of $0\%$ on FRLL with a mean MACER of $90.9\%$ and $98.1\%$, respectively. For DifFIQA(R), the mean bona fide score is $0.86$ on FEI and $0.93$ on FRLL. The datasets differ in capture setup, image resolution, and subjects, and our experiments do not isolate the cause of the shift. The threshold is thus independent of the attacks, as required by (iii), but depends on the bona fide dataset.

Limitations: the bona fide sets are small (204 and 200 images), as reflected in the confidence intervals. The operating point of BSCER $=20\%$ follows SYN-MAD 2022 \cite{DBLP:conf/icb/HuberBLRRDNGSCT22} to allow comparison with that benchmark and is not intended as a setting for deployed systems. The MorDIFF and Arc2Morph attacks are our reproductions with the authors' code, and the FEI sets are used as provided.

\section{Conclusion}
\label{sec:conclusion}
We evaluated face image quality as a training-free signal for single-image MAD, using 17 FIQA methods on 20 morphing attack sets from two datasets, with the score direction of each method fixed in advance. On FRLL, the approach is effective: without morph data or training, DifFIQA(R) reaches a mean D-EER of $11.42\%$ and a mean MACER of $10.75\%$. However, a strong response to morphing does not imply good detection performance, since the shift can be opposite to the meaning of the score. On the diffusion-based attacks, up to 11 of 17 methods assign higher quality to morphs, and on DCMorph the method with the largest reversed separation is the worst detector, while another method detects this attack best. Reporting only the unsigned FDR, or selecting the direction per attack, would present such reversals as successful detection. Three further findings follow. First, the mean performance over attacks hides large differences between attacks: GraFIQs(L), the best detector of DCMorph and Arc2Morph on FRLL, performs poorly on the landmark-based and GAN-based attacks. Second, the FIQA performance of a method predicts its MAD performance only moderately and not robustly. Third, detection performance differs strongly between the datasets, with a lowest mean D-EER of $39.75\%$ on FEI, and a threshold set on the bona fide data of one dataset does not transfer to the other. Future work should investigate whether a threshold calibrated on held-out bona fide images of the target domain generalizes to new images of that domain, and which properties of a FIQA method determine the direction of its response to morphing. Since FIQA usually precedes MAD to reject low-quality samples, trained MAD detectors could also be evaluated with and without such a rejection, on the retained samples.

\section*{Acknowledgment}
This research work has been funded by the German Federal Ministry of Education and Research and the Hessen State Ministry for Higher Education, Research and the Arts within their joint support of the National Research Center for Applied Cybersecurity ATHENE.

\bibliographystyle{IEEEtran}
\bibliography{IEEEtranControl,references}

\clearpage
\onecolumn
\appendices
\newcommand{\fitwidth}[1]{\resizebox{\ifdim\width>\linewidth\linewidth\else\width\fi}{!}{#1}}

The appendices report further results of the experiments in the main paper. All values are computed from the same scores, images, and metric definitions (Sec.~\ref{sec:methodology}). The face detector of SelfMAD++ finds no face in 10 FRLL Arc2Morph images and in 1 LADIMO-inspired image. These images are excluded for all methods, so that all methods are evaluated on the same images.

\FloatBarrier
\section{Example Images}
\label{app:examples}
\begin{figure}[!htbp]
\centering
\includegraphics[width=\textwidth]{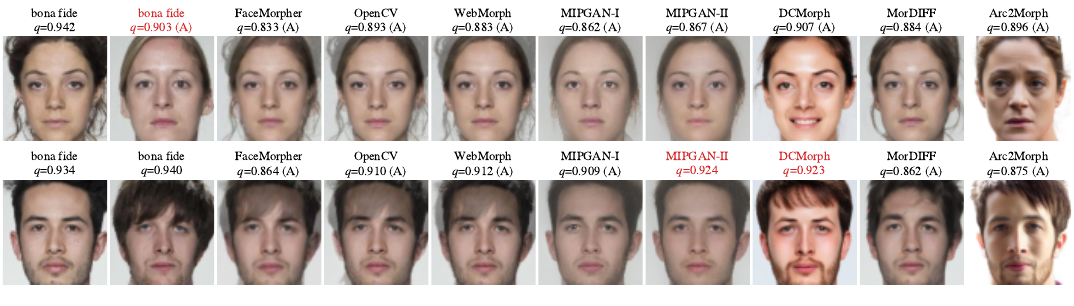}
\caption{Two FRLL morphing pairs: bona fide images and the morph of each attack set, as aligned $112\times112$ inputs, with the DifFIQA(R) score $q$. (A) marks a classification as attack at BSCER $=20\%$, and red marks a classification error. Bona fide images: Face Research Lab London Set \cite{FRLL}, CC BY 4.0.}
\label{fig:app_examples_frll}
\end{figure}

Fig.~\ref{fig:app_examples_frll} shows two FRLL morphing pairs. Each row shows the two bona fide images of a pair, taken from the Face Research Lab London Set \cite{FRLL}, followed by the morph of this pair from each attack set. All images are the aligned $112\times112$ inputs that the methods receive (Sec.~\ref{sec:experimental}). The two pairs were chosen by file name, not by their scores. Each image is labeled with its DifFIQA(R) score $q$. (A) marks an image classified as attack at the FRLL threshold $\tau_m=0.922$, and a red label marks a wrong decision.

\FloatBarrier
\section{MACER per Attack}
\label{app:macer}
\begin{figure}[!htbp]
\centering
\includegraphics[width=\textwidth]{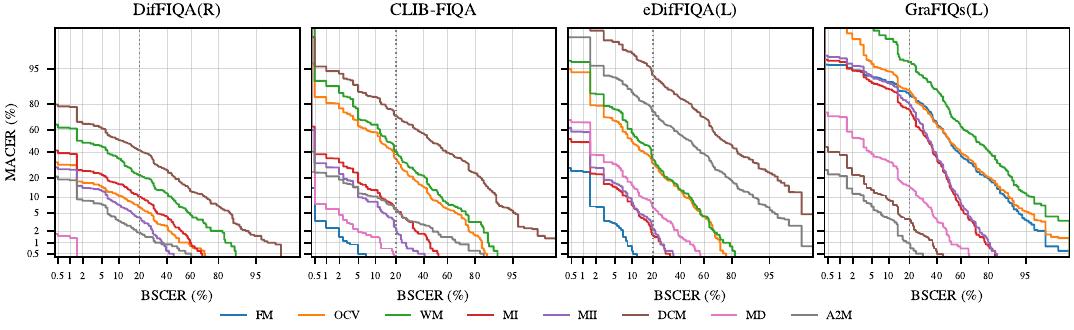}
\caption{DET curves (MACER over BSCER, normal deviate scale) on the FRLL attacks for the four methods with the lowest mean D-EER on FRLL. Dotted: operating point BSCER $=20\%$. Attack abbreviations as in Tab.~\ref{tab:mad_results}.}
\label{fig:app_det}
\end{figure}

\begin{figure}[!htbp]
\centering
\includegraphics[width=\textwidth]{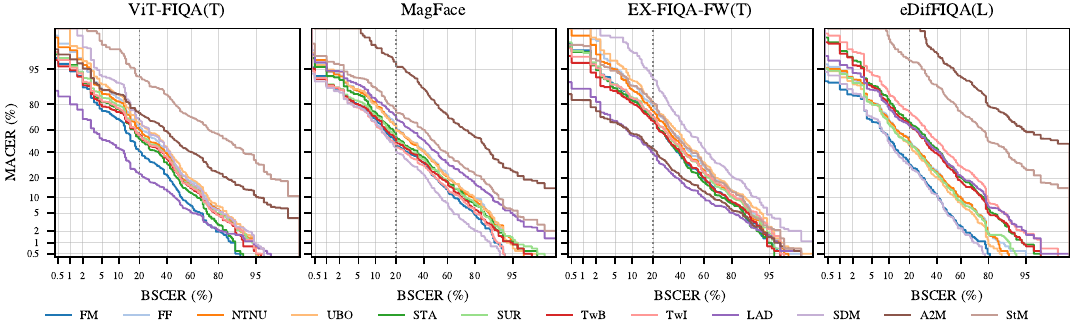}
\caption{DET curves on the FEI attacks for the four methods with the lowest mean D-EER on FEI, as in Fig.~\ref{fig:app_det}. Attack abbreviations as in Tab.~\ref{tab:fei}.}
\label{fig:app_det_fei}
\end{figure}

Tabs.~\ref{tab:mad_results} and~\ref{tab:fei} report the mean MACER over the attacks of each dataset. Tabs.~\ref{tab:app_macer_frll} and~\ref{tab:app_macer_fei} list the MACER of each attack at the same operating point, i.e.\ with the threshold $\tau_m$ set at BSCER $=20\%$ on the bona fide images of the dataset. At this operating point, a random score yields a MACER of about $80\%$. A higher MACER means that morphs pass the threshold more often than bona fide images. This is the case for 52 of the 136 method-attack pairs on FRLL and 74 of the 204 pairs on FEI. DifFIQA(R) reaches a MACER of $0.00\%$ on FaceMorpher, and the lowest DCMorph MACER is reached by GraFIQs(L) ($3.70\%$). Fig.~\ref{fig:app_det} shows the detection error tradeoff (DET) curves, i.e.\ MACER over BSCER for all thresholds, of the four methods with the lowest mean D-EER on FRLL. As in Tab.~\ref{tab:mad_results}, DCMorph is the hardest attack for DifFIQA(R), CLIB-FIQA, and eDifFIQA(L), while GraFIQs(L) detects Arc2Morph and DCMorph best. Fig.~\ref{fig:app_det_fei} shows the same for the four methods with the lowest mean D-EER on FEI.

\begin{table}[!htbp]
\centering
\caption{MACER (\%) per attack on FRLL at the fixed bona fide operating point (BSCER $=20\%$, Sec.~\ref{sec:methodology}). The last column is the mean MACER of Tab.~\ref{tab:mad_results}. Bold marks the lowest value of a FIQA method per column. Attack abbreviations as in Tab.~\ref{tab:mad_results}.}
\label{tab:app_macer_frll}
\footnotesize
\setlength{\tabcolsep}{3pt}
\fitwidth{%
\begin{tabular}{@{}lccccccccc@{}}
\toprule
\textbf{Method} & \textbf{FM} & \textbf{OCV} & \textbf{WM} & \textbf{MI} & \textbf{MII} & \textbf{DCM} & \textbf{MD} & \textbf{A2M} & \textbf{Mean MACER} \\
\midrule
\multicolumn{10}{@{}l}{\textit{Reference only, not for comparison: trained detector on $112\times112$ inputs}} \\
SelfMAD++ \cite{DBLP:journals/inffus/IvanovskaTPS26} & 0.00 & 0.00 & 3.20 & 3.40 & 0.10 & 69.90 & 0.20 & 42.53 & 14.92 \\
\midrule
\multicolumn{10}{@{}l}{\textit{Supervised}} \\
FaceQnet & 78.20 & 86.79 & 80.20 & 83.90 & 83.98 & 80.00 & 86.80 & 70.10 & 81.25 \\
SDD-FIQA & 55.30 & 81.61 & 62.60 & 47.80 & 43.04 & 95.70 & 95.80 & 82.32 & 70.52 \\
CLIB-FIQA & 0.20 & 34.15 & 40.00 & 5.90 & \textbf{2.30} & 71.40 & 0.40 & 5.86 & 20.03 \\
\midrule
\multicolumn{10}{@{}l}{\textit{Unsupervised, non-FR-model}} \\
DifFIQA(R) & \textbf{0.00} & \textbf{6.61} & \textbf{21.80} & 10.20 & 4.00 & 41.60 & \textbf{0.00} & 1.82 & \textbf{10.75} \\
eDifFIQA(L) & 0.30 & 30.59 & 34.20 & \textbf{1.60} & \textbf{2.30} & 93.30 & 6.70 & 74.75 & 30.47 \\
\midrule
\multicolumn{10}{@{}l}{\textit{Unsupervised, FR-probing}} \\
SER-FIQ & 79.50 & 87.80 & 89.60 & 66.20 & 68.47 & 85.10 & 75.80 & 90.91 & 80.42 \\
GraFIQs(L) & 85.70 & 87.40 & 96.40 & 76.20 & 80.08 & \textbf{3.70} & 14.80 & \textbf{1.01} & 55.66 \\
FROQ & 46.20 & 81.81 & 83.20 & 40.20 & 34.93 & 99.90 & 99.30 & 100.00 & 73.19 \\
ViTNT-FIQA & 86.90 & 96.65 & 96.60 & 94.30 & 90.59 & 62.60 & 90.50 & 25.05 & 80.40 \\
PreFIQs & 64.30 & 74.49 & 78.40 & 36.30 & 36.54 & 67.40 & 60.90 & 76.36 & 61.84 \\
ATTN-FIQA & 61.90 & 85.16 & 84.00 & 80.90 & 67.07 & 87.50 & 92.40 & 75.56 & 79.31 \\
\midrule
\multicolumn{10}{@{}l}{\textit{Self-supervised (FR-integrated)}} \\
MagFace & 37.00 & 71.44 & 66.20 & 20.00 & 18.72 & 90.70 & 71.70 & 99.80 & 59.44 \\
PFE & 36.50 & 68.29 & 48.40 & 38.80 & 27.73 & 98.40 & 82.50 & 94.14 & 61.85 \\
CR-FIQA(L) & 43.90 & 71.44 & 49.20 & 10.90 & 11.61 & 90.10 & 90.30 & 96.77 & 58.03 \\
ViT-FIQA(T) & 46.90 & 75.71 & 49.80 & 40.60 & 31.53 & 98.20 & 93.40 & 89.09 & 65.65 \\
CARPM-FIQA(L) & 45.40 & 70.33 & 45.80 & 12.80 & 6.91 & 91.10 & 89.20 & 93.94 & 56.93 \\
EX-FIQA-FW(T) & 71.10 & 91.46 & 82.40 & 78.40 & 69.37 & 97.80 & 98.20 & 70.20 & 82.37 \\
\bottomrule
\end{tabular}}
\end{table}

\begin{table}[!htbp]
\centering
\caption{MACER (\%) per attack on FEI at the fixed bona fide operating point (BSCER $=20\%$). The last column is the mean MACER of Tab.~\ref{tab:fei}. Bold as in Tab.~\ref{tab:app_macer_frll}, attack abbreviations as in Tab.~\ref{tab:fei}.}
\label{tab:app_macer_fei}
\scriptsize
\setlength{\tabcolsep}{3pt}
\fitwidth{%
\begin{tabular}{@{}lccccccccccccc@{}}
\toprule
\textbf{Method} & \textbf{FM} & \textbf{FF} & \textbf{NTNU} & \textbf{UBO} & \textbf{STA} & \textbf{SUR} & \textbf{TwB} & \textbf{TwI} & \textbf{LAD} & \textbf{SDM} & \textbf{A2M} & \textbf{StM} & \textbf{Mean MACER} \\
\midrule
\multicolumn{14}{@{}l}{\textit{Reference only, not for comparison: trained detector on $112\times112$ inputs}} \\
SelfMAD++ \cite{DBLP:journals/inffus/IvanovskaTPS26} & 0.0 & 18.6 & 38.7 & 17.7 & 70.5 & 13.6 & 55.5 & 84.6 & 6.6 & 8.5 & 99.3 & 13.2 & 35.57 \\
\midrule
\multicolumn{14}{@{}l}{\textit{Supervised}} \\
FaceQnet & 81.9 & 80.6 & 80.7 & 80.6 & 82.0 & 80.0 & 81.8 & 79.6 & 64.3 & 86.3 & 74.0 & 70.6 & 78.54 \\
SDD-FIQA & 78.3 & 80.4 & 80.8 & 81.1 & 80.8 & 81.0 & 77.2 & 75.6 & 54.2 & 89.7 & 80.2 & 83.0 & 78.53 \\
CLIB-FIQA & 40.0 & 61.6 & 57.2 & 60.0 & 79.6 & 63.6 & 73.4 & 85.0 & 71.2 & 76.4 & 99.8 & 86.4 & 71.19 \\
\midrule
\multicolumn{14}{@{}l}{\textit{Unsupervised, non-FR-model}} \\
DifFIQA(R) & \textbf{26.9} & 52.5 & 55.4 & 53.8 & 69.4 & 60.5 & 68.5 & 86.6 & 81.0 & 39.9 & 100.0 & 97.9 & 66.03 \\
eDifFIQA(L) & 30.4 & \textbf{50.7} & \textbf{50.5} & \textbf{46.4} & 66.9 & \textbf{48.2} & 65.3 & 75.1 & 64.0 & \textbf{29.1} & 99.9 & 96.8 & 60.28 \\
\midrule
\multicolumn{14}{@{}l}{\textit{Unsupervised, FR-probing}} \\
SER-FIQ & 81.1 & 86.0 & 86.1 & 85.4 & 85.9 & 87.6 & 83.3 & 83.1 & 95.4 & 69.1 & 79.9 & 77.8 & 83.39 \\
GraFIQs(L) & 82.5 & 90.0 & 80.5 & 79.9 & 90.3 & 79.2 & 86.6 & 91.8 & 87.9 & 73.4 & 25.6 & 24.9 & 74.38 \\
FROQ & 65.7 & 77.2 & 80.7 & 79.5 & 76.4 & 79.9 & 76.8 & 80.2 & 63.8 & 96.6 & 99.6 & 83.9 & 80.03 \\
ViTNT-FIQA & 95.5 & 95.0 & 93.5 & 95.0 & 90.3 & 90.1 & 88.1 & 84.7 & 70.8 & 97.8 & \textbf{0.9} & \textbf{21.2} & 76.91 \\
PreFIQs & 74.6 & 78.7 & 79.5 & 78.6 & 77.6 & 75.7 & 77.0 & 76.6 & 90.0 & 62.6 & 87.9 & 83.8 & 78.55 \\
ATTN-FIQA & 87.6 & 91.4 & 88.0 & 92.1 & 82.5 & 80.5 & 82.1 & 79.4 & 47.8 & 92.6 & 65.3 & 63.6 & 79.41 \\
\midrule
\multicolumn{14}{@{}l}{\textit{Self-supervised (FR-integrated)}} \\
MagFace & 49.9 & 65.1 & 63.8 & 64.2 & 57.2 & 54.5 & \textbf{52.2} & \textbf{47.3} & 71.7 & 43.9 & 96.5 & 76.6 & 61.91 \\
PFE & 39.8 & 58.8 & 55.4 & 58.9 & 57.0 & 59.0 & 54.8 & 57.0 & 34.5 & 77.4 & 89.1 & 95.1 & 61.40 \\
CR-FIQA(L) & 62.6 & 75.3 & 74.8 & 76.8 & 71.5 & 68.7 & 70.4 & 64.4 & 46.6 & 71.5 & 92.8 & 79.4 & 71.23 \\
ViT-FIQA(T) & 40.9 & 62.8 & 60.5 & 65.7 & \textbf{52.2} & 56.3 & 55.9 & 55.0 & \textbf{22.8} & 68.5 & 73.9 & 93.0 & \textbf{58.96} \\
CARPM-FIQA(L) & 56.7 & 72.1 & 72.2 & 73.3 & 70.5 & 66.9 & 69.2 & 66.8 & 52.6 & 66.0 & 86.3 & 79.1 & 69.31 \\
EX-FIQA-FW(T) & 69.8 & 79.7 & 76.0 & 81.9 & 70.5 & 71.2 & 68.1 & 72.0 & 39.7 & 93.0 & 43.5 & 81.4 & 70.56 \\
\bottomrule
\end{tabular}}
\end{table}

\FloatBarrier
\section{Confidence Intervals}
\label{app:ci}
Tab.~\ref{tab:app_ci} reports the $95\%$ confidence interval of the mean D-EER for every method, computed with the bootstrap of Sec.~\ref{sec:methodology}. On FRLL, the interval of DifFIQA(R) lies below that of the second-ranked method, CLIB-FIQA ([14.7, 19.7]). On FEI, the interval of ViT-FIQA(T) overlaps those of 7 other methods, so the best methods on FEI cannot be ranked reliably.

\begin{table}[!htbp]
\centering
\caption{Mean D-EER (\%) with $95\%$ identity-level bootstrap confidence intervals (1,000 replicates, Sec.~\ref{sec:methodology}).}
\label{tab:app_ci}
\footnotesize
\setlength{\tabcolsep}{4pt}
\begin{tabular}{@{}lcc@{}}
\toprule
\textbf{Method} & \textbf{FRLL} & \textbf{FEI} \\
\midrule
\multicolumn{3}{@{}l}{\textit{Supervised}} \\
FaceQnet & 48.39 [45.2, 51.9] & 47.76 [44.7, 51.1] \\
SDD-FIQA & 47.03 [44.2, 50.0] & 47.02 [43.7, 50.1] \\
CLIB-FIQA & 16.99 [14.7, 19.7] & 44.82 [42.1, 47.7] \\
\midrule
\multicolumn{3}{@{}l}{\textit{Unsupervised, non-FR-model}} \\
DifFIQA(R) & 11.42 [9.0, 14.0] & 44.25 [42.2, 46.9] \\
eDifFIQA(L) & 24.34 [22.3, 27.0] & 40.84 [38.4, 43.5] \\
\midrule
\multicolumn{3}{@{}l}{\textit{Unsupervised, FR-probing}} \\
SER-FIQ & 48.18 [45.0, 51.0] & 52.74 [49.7, 55.1] \\
GraFIQs(L) & 33.85 [31.1, 37.2] & 45.97 [43.6, 48.6] \\
FROQ & 56.44 [54.2, 58.5] & 50.46 [47.6, 53.3] \\
ViTNT-FIQA & 56.55 [52.8, 59.2] & 52.56 [49.9, 54.8] \\
PreFIQs & 42.70 [39.7, 45.4] & 49.39 [46.1, 52.2] \\
ATTN-FIQA & 50.45 [46.5, 53.9] & 48.02 [44.8, 51.2] \\
\midrule
\multicolumn{3}{@{}l}{\textit{Self-supervised (FR-integrated)}} \\
MagFace & 41.87 [38.3, 44.7] & 40.55 [37.5, 43.8] \\
PFE & 42.03 [39.4, 45.1] & 41.45 [38.7, 43.9] \\
CR-FIQA(L) & 43.22 [40.6, 46.4] & 46.06 [43.4, 48.9] \\
ViT-FIQA(T) & 43.65 [41.1, 46.1] & 39.75 [37.3, 42.6] \\
CARPM-FIQA(L) & 41.08 [38.9, 43.8] & 43.32 [40.2, 45.9] \\
EX-FIQA-FW(T) & 49.49 [46.6, 52.0] & 40.83 [38.2, 43.3] \\
\bottomrule
\end{tabular}
\end{table}

\FloatBarrier
\section{Selecting the Score Direction}
\label{app:selection}
Our protocol fixes the score direction of each method in advance (Sec.~\ref{sec:methodology}). Tab.~\ref{tab:app_selection} shows how the mean D-EER would change if the direction were chosen using the attack labels, either once per method, as in \cite{DBLP:journals/iet-bmt/FuD22}, or separately for each attack. Both choices violate condition (iii) of Sec.~\ref{sec:methodology} and are shown only for comparison. The per-method choice inverts FROQ, ViTNT-FIQA, and ATTN-FIQA on FRLL and SER-FIQ, FROQ, and ViTNT-FIQA on FEI. The per-attack choice helps FROQ most, lowering its mean D-EER on FRLL from $56.44\%$ to $25.93\%$, although FROQ is the worst detector of DCMorph with the fixed direction (Sec.~\ref{sec:mad_results}). The three best methods on FRLL stay the same in all three settings. On FEI, the per-attack choice makes eDifFIQA(L) ($32.77\%$) the best method instead of ViT-FIQA(T).

\begin{table}[!htbp]
\centering
\caption{Mean D-EER (\%) with the fixed score direction of our protocol (Fixed) and with a direction selected using the attack labels, per method as in \cite{DBLP:journals/iet-bmt/FuD22} (Method) or per method and attack (Attack). $^{\ast}$: the per-method selection inverts the score. Both selections use attack data, are not training-free, and are reported for comparison only.}
\label{tab:app_selection}
\footnotesize
\setlength{\tabcolsep}{3pt}
\fitwidth{%
\begin{tabular}{@{}lcccccc@{}}
\toprule
& \multicolumn{3}{c}{\textbf{FRLL}} & \multicolumn{3}{c}{\textbf{FEI}} \\
\cmidrule(lr){2-4}\cmidrule(lr){5-7}
\textbf{Method} & \textbf{Fixed} & \textbf{Method} & \textbf{Attack} & \textbf{Fixed} & \textbf{Method} & \textbf{Attack} \\
\midrule
\multicolumn{7}{@{}l}{\textit{Supervised}} \\
FaceQnet & 48.39 & 48.39 & 47.60 & 47.76 & 47.76 & 46.97 \\
SDD-FIQA & 47.03 & 47.03 & 37.98 & 47.02 & 47.02 & 46.69 \\
CLIB-FIQA & 16.99 & 16.99 & 16.99 & 44.82 & 44.82 & 37.41 \\
\midrule
\multicolumn{7}{@{}l}{\textit{Unsupervised, non-FR-model}} \\
DifFIQA(R) & 11.42 & 11.42 & 11.42 & 44.25 & 44.25 & 33.12 \\
eDifFIQA(L) & 24.34 & 24.34 & 21.04 & 40.84 & 40.84 & 32.77 \\
\midrule
\multicolumn{7}{@{}l}{\textit{Unsupervised, FR-probing}} \\
SER-FIQ & 48.18 & 48.18 & 42.50 & 52.74 & 47.22$^{\ast}$ & 45.64 \\
GraFIQs(L) & 33.85 & 33.85 & 30.77 & 45.97 & 45.97 & 42.08 \\
FROQ & 56.44 & 43.56$^{\ast}$ & 25.93 & 50.46 & 49.54$^{\ast}$ & 42.21 \\
ViTNT-FIQA & 56.55 & 43.45$^{\ast}$ & 32.95 & 52.56 & 47.44$^{\ast}$ & 35.44 \\
PreFIQs & 42.70 & 42.70 & 41.90 & 49.39 & 49.39 & 46.32 \\
ATTN-FIQA & 50.45 & 49.55$^{\ast}$ & 44.23 & 48.02 & 48.02 & 43.59 \\
\midrule
\multicolumn{7}{@{}l}{\textit{Self-supervised (FR-integrated)}} \\
MagFace & 41.87 & 41.87 & 31.85 & 40.55 & 40.55 & 37.88 \\
PFE & 42.03 & 42.03 & 32.63 & 41.45 & 41.45 & 35.20 \\
CR-FIQA(L) & 43.22 & 43.22 & 31.02 & 46.06 & 46.06 & 43.02 \\
ViT-FIQA(T) & 43.65 & 43.65 & 31.73 & 39.75 & 39.75 & 36.84 \\
CARPM-FIQA(L) & 41.08 & 41.08 & 30.97 & 43.32 & 43.32 & 41.74 \\
EX-FIQA-FW(T) & 49.49 & 49.49 & 39.32 & 40.83 & 40.83 & 40.25 \\
\bottomrule
\end{tabular}}
\end{table}

\FloatBarrier
\section{Threshold Transfer for All Methods}
\label{app:transfer}
\begin{figure}[!htbp]
\centering
\includegraphics[width=\textwidth]{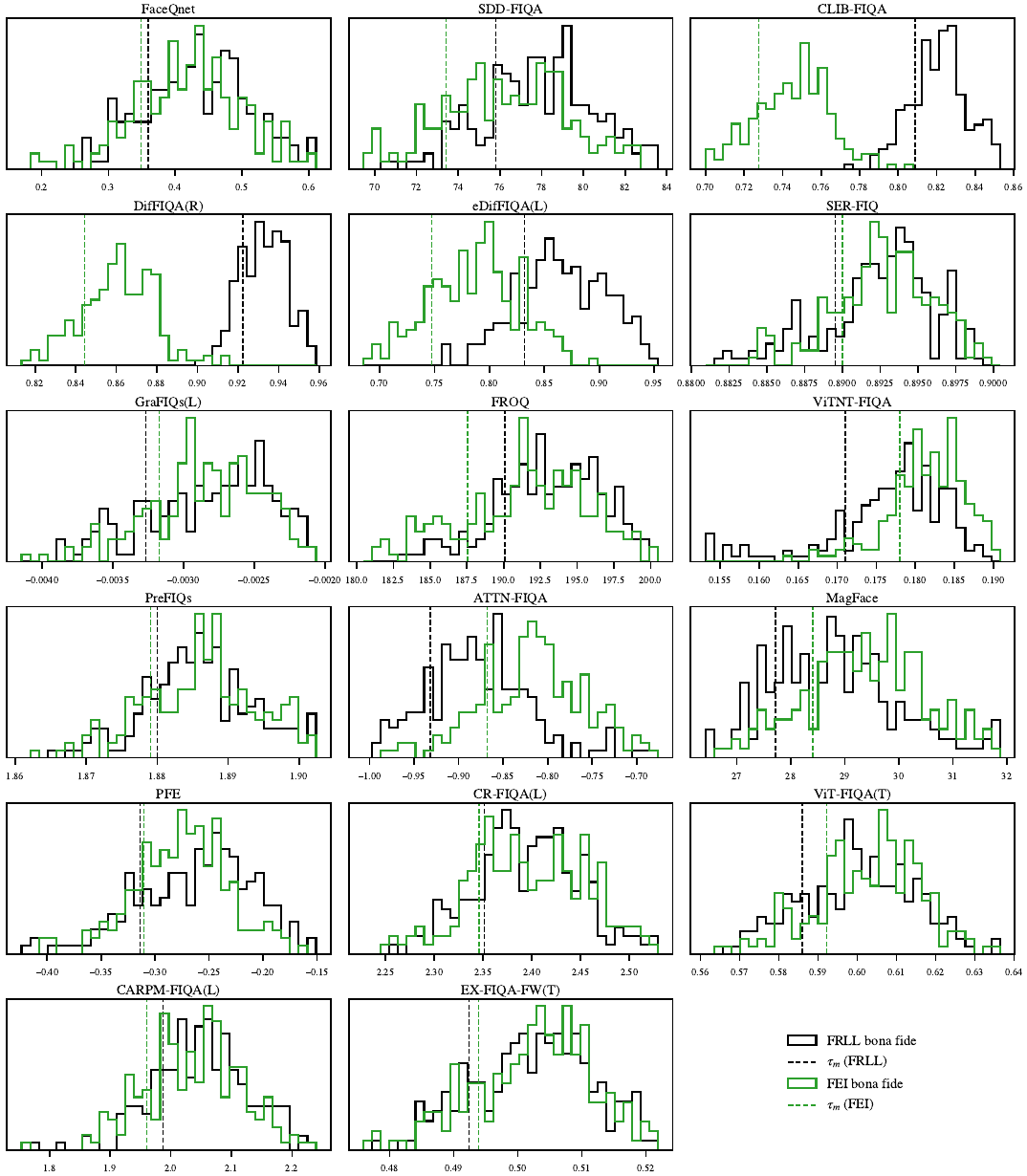}
\caption{Distribution of the direction-normalized score $q_m$ of every method on the FRLL and FEI bona fide images, with the threshold $\tau_m$ at BSCER $=20\%$ set on each dataset (dashed).}
\label{fig:app_bfshift}
\end{figure}

Sec.~\ref{sec:transfer} reports the threshold transfer for DifFIQA(R) and CLIB-FIQA, and Tab.~\ref{tab:app_transfer} lists it for all methods. For the three best methods on FRLL, the FRLL threshold yields a BSCER of $86.0$-$100.0\%$ on FEI, and the FEI threshold yields a BSCER of $0\%$ on FRLL with a mean MACER of $80.0$-$98.1\%$. For FaceQnet, SER-FIQ, PreFIQs, PFE, CR-FIQA(L), and EX-FIQA-FW(T), the BSCER stays between $15\%$ and $25\%$ in both directions, but these methods detect morphs poorly on both datasets, with mean D-EERs of at least $40.83\%$. Fig.~\ref{fig:app_bfshift} shows the bona fide scores of all methods on both datasets, together with the threshold set on each. For the three best methods on FRLL, most FEI bona fide scores lie below the FRLL threshold.

\begin{table}[!htbp]
\centering
\caption{Threshold transfer between the datasets. $\tau_m$ is set at BSCER $=20\%$ on the source dataset and applied unchanged to the target dataset. BSCER and mean MACER (\%) are measured on the target. In-dataset is the mean MACER with $\tau_m$ set on the target dataset itself (Tabs.~\ref{tab:mad_results} and~\ref{tab:fei}).}
\label{tab:app_transfer}
\footnotesize
\setlength{\tabcolsep}{3pt}
\fitwidth{%
\begin{tabular}{@{}lcccccc@{}}
\toprule
& \multicolumn{3}{c}{\textbf{FRLL $\rightarrow$ FEI}} & \multicolumn{3}{c}{\textbf{FEI $\rightarrow$ FRLL}} \\
\cmidrule(lr){2-4}\cmidrule(lr){5-7}
\textbf{Method} & \textbf{BSCER} & \textbf{MACER} & \textbf{In-dataset} & \textbf{BSCER} & \textbf{MACER} & \textbf{In-dataset} \\
\midrule
\multicolumn{7}{@{}l}{\textit{Supervised}} \\
FaceQnet & 23.5 & 74.7 & 78.5 & 17.6 & 84.7 & 81.2 \\
SDD-FIQA & 43.5 & 49.5 & 78.5 & 4.4 & 90.7 & 70.5 \\
CLIB-FIQA & 100.0 & 2.9 & 71.2 & 0.0 & 98.1 & 20.0 \\
\midrule
\multicolumn{7}{@{}l}{\textit{Unsupervised, non-FR-model}} \\
DifFIQA(R) & 100.0 & 0.8 & 66.0 & 0.0 & 90.9 & 10.8 \\
eDifFIQA(L) & 86.0 & 12.1 & 60.3 & 0.0 & 80.0 & 30.5 \\
\midrule
\multicolumn{7}{@{}l}{\textit{Unsupervised, FR-probing}} \\
SER-FIQ & 17.5 & 86.4 & 83.4 & 21.6 & 76.3 & 80.4 \\
GraFIQs(L) & 14.5 & 79.8 & 74.4 & 24.0 & 51.9 & 55.7 \\
FROQ & 33.0 & 64.7 & 80.0 & 8.3 & 87.7 & 73.2 \\
ViTNT-FIQA & 3.5 & 91.4 & 76.9 & 51.5 & 58.8 & 80.4 \\
PreFIQs & 23.0 & 75.5 & 78.5 & 17.2 & 65.7 & 61.8 \\
ATTN-FIQA & 3.5 & 95.5 & 79.4 & 58.8 & 41.9 & 79.3 \\
\midrule
\multicolumn{7}{@{}l}{\textit{Self-supervised (FR-integrated)}} \\
MagFace & 8.5 & 79.7 & 61.9 & 39.7 & 43.5 & 59.4 \\
PFE & 18.0 & 63.7 & 61.4 & 20.6 & 60.2 & 61.8 \\
CR-FIQA(L) & 23.0 & 68.3 & 71.2 & 17.6 & 60.1 & 58.0 \\
ViT-FIQA(T) & 13.0 & 72.3 & 59.0 & 28.9 & 53.2 & 65.7 \\
CARPM-FIQA(L) & 27.0 & 58.1 & 69.3 & 11.8 & 64.8 & 56.9 \\
EX-FIQA-FW(T) & 16.5 & 74.9 & 70.6 & 23.0 & 78.3 & 82.4 \\
\bottomrule
\end{tabular}}
\end{table}

\FloatBarrier
\section{FIQA Ranking and MAD Ranking}
\label{app:utility}
\begin{figure}[!htbp]
\centering
\includegraphics[width=0.85\textwidth]{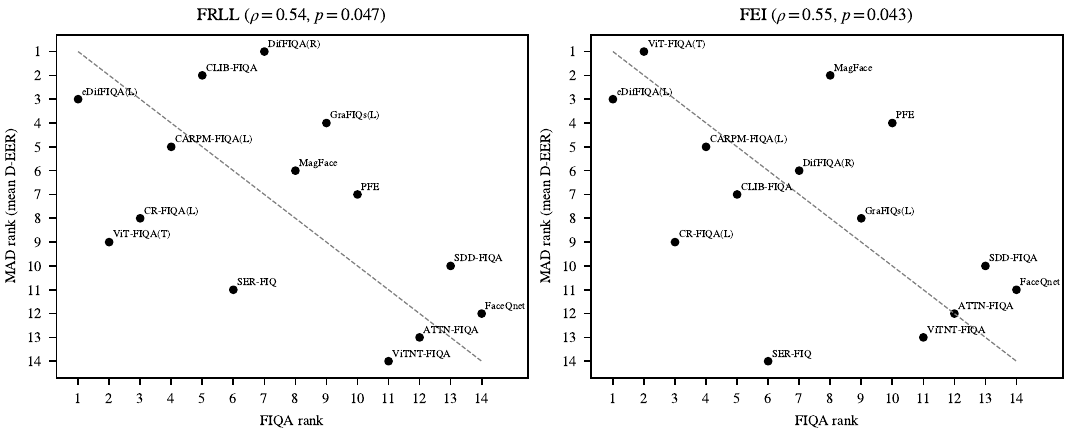}
\caption{FIQA rank over MAD rank (mean D-EER) of the 14 methods with a FIQA evaluation, with the Spearman correlation of Sec.~\ref{sec:mad_results}. Dashed: identical ranks.}
\label{fig:app_rank}
\end{figure}

Tab.~\ref{tab:app_utility} details the comparison of Sec.~\ref{sec:mad_results}. The FIQA evaluation uses the FR models ArcFace, CurricularFace, ElasticFace, and MagFace, the datasets Adience, AgeDB-30, CFP-FP, LFW, CALFW, CPLFW, XQLFW, and IJB-C, and false match rates of $10^{-3}$ and $10^{-4}$. eDifFIQA(L) ranks first by FIQA performance and third by mean D-EER on both datasets. The largest difference, eight positions, is for SER-FIQ, which ranks 6th by FIQA performance and 14th by mean D-EER on FEI. Fig.~\ref{fig:app_rank} plots both rankings against each other.

\begin{table}[!htbp]
\centering
\caption{FIQA ranking and MAD ranking of the 14 methods with a FIQA evaluation (Sec.~\ref{sec:mad_results}). Avg.: average rank by the partial area under the error-versus-discard curve over the 64 combinations. FIQA is the resulting rank, MAD the rank by mean D-EER on FRLL and FEI, and $p_{-m}$ the $p$-value of the Spearman correlation after removing the method.}
\label{tab:app_utility}
\footnotesize
\setlength{\tabcolsep}{3pt}
\fitwidth{%
\begin{tabular}{@{}lcccccc@{}}
\toprule
& & & \multicolumn{2}{c}{\textbf{MAD}} & \multicolumn{2}{c}{$\boldsymbol{p_{-m}}$} \\
\cmidrule(lr){4-5}\cmidrule(lr){6-7}
\textbf{Method} & \textbf{Avg.} & \textbf{FIQA} & \textbf{FRLL} & \textbf{FEI} & \textbf{FRLL} & \textbf{FEI} \\
\midrule
\multicolumn{7}{@{}l}{\textit{Supervised}} \\
FaceQnet & 13.62 & 14 & 12 & 11 & 0.122 & 0.078 \\
SDD-FIQA & 11.19 & 13 & 10 & 10 & 0.090 & 0.078 \\
CLIB-FIQA & 4.98 & 5 & 2 & 7 & 0.052 & 0.058 \\
\midrule
\multicolumn{7}{@{}l}{\textit{Unsupervised, non-FR-model}} \\
DifFIQA(R) & 6.66 & 7 & 1 & 6 & 0.031 & 0.039 \\
eDifFIQA(L) & 3.25 & 1 & 3 & 3 & 0.098 & 0.098 \\
\midrule
\multicolumn{7}{@{}l}{\textit{Unsupervised, FR-probing}} \\
SER-FIQ & 6.55 & 6 & 11 & 14 & 0.037 & 0.010 \\
GraFIQs(L) & 7.97 & 9 & 4 & 8 & 0.044 & 0.058 \\
ViTNT-FIQA & 9.19 & 11 & 14 & 13 & 0.112 & 0.074 \\
ATTN-FIQA & 10.78 & 12 & 13 & 12 & 0.112 & 0.074 \\
\midrule
\multicolumn{7}{@{}l}{\textit{Self-supervised (FR-integrated)}} \\
MagFace & 7.33 & 8 & 6 & 2 & 0.055 & 0.025 \\
PFE & 9.08 & 10 & 7 & 4 & 0.061 & 0.031 \\
CR-FIQA(L) & 4.77 & 3 & 8 & 9 & 0.027 & 0.039 \\
ViT-FIQA(T) & 4.73 & 2 & 9 & 1 & 0.027 & 0.128 \\
CARPM-FIQA(L) & 4.91 & 4 & 5 & 5 & 0.049 & 0.064 \\
\bottomrule
\end{tabular}}
\end{table}

\FloatBarrier
\section{Per-Attack Summary}
\label{app:summary}
\begin{figure}[!htbp]
\centering
\includegraphics[width=\textwidth]{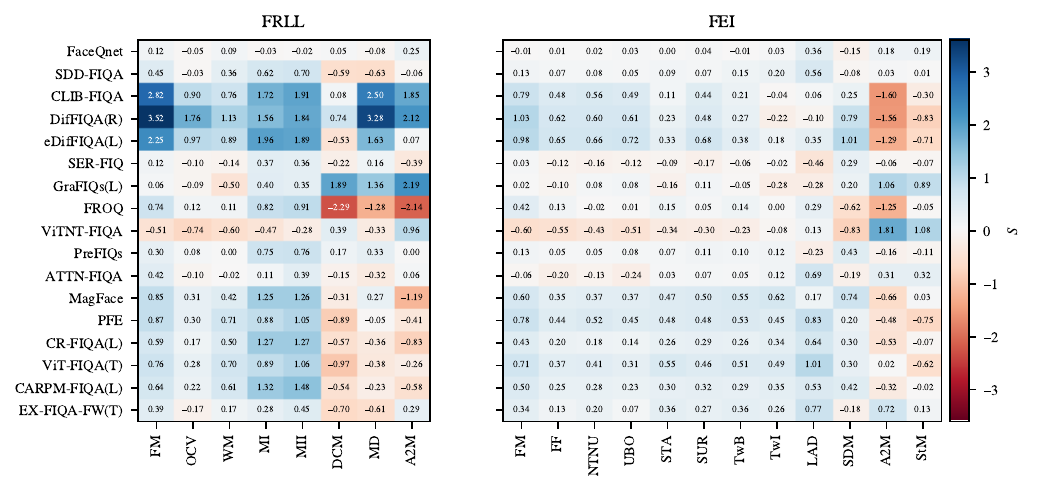}
\caption{Signed separation $S$ (Eq.~(\ref{eq:fdr})) of all methods and attacks on FRLL and FEI. Blue marks morphs with lower quality than bona fide images (expected direction), and red marks a reversal. Attack abbreviations as in Tabs.~\ref{tab:mad_results} and~\ref{tab:fei}.}
\label{fig:app_heatmap}
\end{figure}

\begin{figure}[!htbp]
\centering
\includegraphics[width=\textwidth]{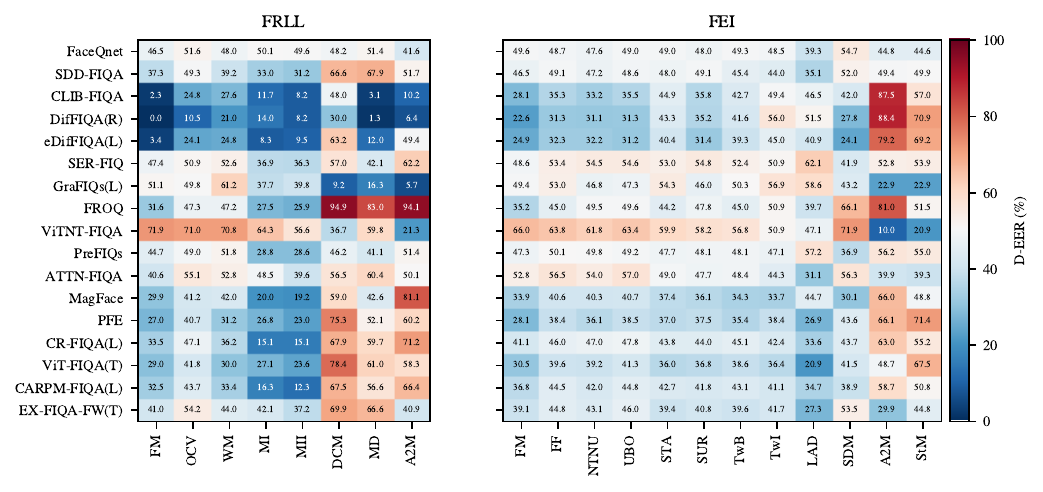}
\caption{D-EER (\%) of all methods and attacks on FRLL and FEI (Tabs.~\ref{tab:mad_results} and~\ref{tab:fei}). Blue marks values below $50\%$, and red marks values above $50\%$, i.e.\ a reversed ordering.}
\label{fig:app_deer_heatmap}
\end{figure}

\begin{figure}[!htbp]
\centering
\includegraphics[width=\textwidth]{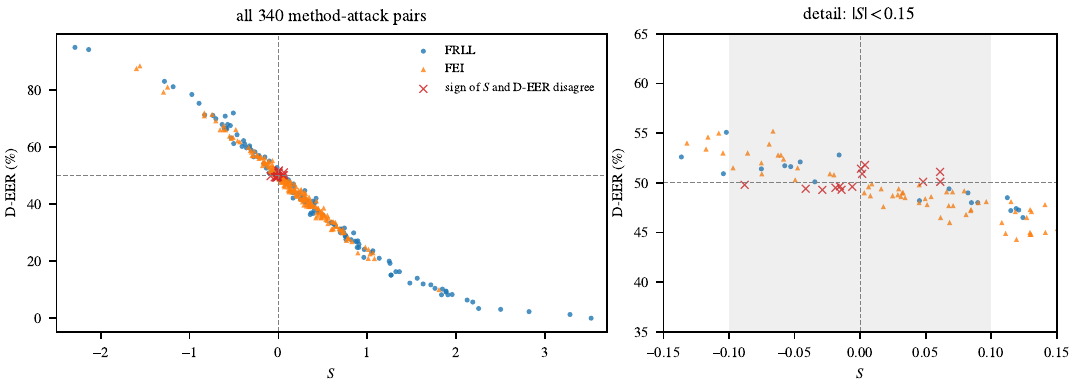}
\caption{Signed separation $S$ against D-EER for all method-attack pairs of FRLL and FEI. Crosses: pairs for which the sign of $S$ and the D-EER disagree. The right panel shows the region $|S|<0.15$, with $|S|<0.1$ shaded.}
\label{fig:app_s_vs_deer}
\end{figure}

Tab.~\ref{tab:app_summary} summarizes each attack over the 17 FIQA methods. The median D-EER exceeds $50\%$ on DCMorph, MorDIFF, and Arc2Morph on FRLL and on Arc2Morph and StableMorph on FEI, i.e.\ on these attacks more than half of the methods perform worse than a random score. Fig.~\ref{fig:app_heatmap} shows the signed separation $S$ of Tabs.~\ref{tab:mad_results} and~\ref{tab:fei} for all methods and attacks, with reversals ($S<0$) in red, and Fig.~\ref{fig:app_deer_heatmap} shows the D-EERs in the same layout. Fig.~\ref{fig:app_s_vs_deer} plots $S$ against the D-EER for all 340 method-attack pairs. As stated in Sec.~\ref{sec:methodology}, the sign of $S$ and the D-EER disagree only for $|S|<0.1$, which is the case for 13 pairs.

\begin{table}[!htbp]
\centering
\caption{Per-attack summary over the 17 FIQA methods: number of reversals ($S<0$), method with the lowest D-EER (\%), and median D-EER (\%).}
\label{tab:app_summary}
\footnotesize
\setlength{\tabcolsep}{3pt}
\begin{tabular}{@{}lccc@{}}
\toprule
\textbf{Attack} & \textbf{Rev.} & \textbf{Lowest D-EER} & \textbf{Median} \\
\midrule
\multicolumn{4}{@{}l}{\textit{FRLL}} \\
FaceMorpher & 1 & DifFIQA(R) (0.00) & 33.50 \\
OpenCV & 7 & DifFIQA(R) (10.47) & 47.26 \\
WebMorph & 4 & DifFIQA(R) (21.00) & 42.00 \\
MIPGAN-I & 2 & eDifFIQA(L) (8.30) & 27.50 \\
MIPGAN-II & 2 & CLIB-FIQA (8.21) & 25.93 \\
DCMorph & 11 & GraFIQs(L) (9.20) & 59.00 \\
MorDIFF & 10 & DifFIQA(R) (1.30) & 52.10 \\
Arc2Morph & 8 & GraFIQs(L) (5.66) & 51.41 \\
\midrule
\multicolumn{4}{@{}l}{\textit{FEI}} \\
FaceMorpher & 3 & DifFIQA(R) (22.6) & 39.1 \\
FaceFusion & 4 & DifFIQA(R) (31.3) & 45.0 \\
NTNU FaceMorph & 4 & DifFIQA(R) (31.1) & 46.8 \\
UBO Morph & 3 & eDifFIQA(L) (31.2) & 47.3 \\
Twente STASM & 3 & ViT-FIQA(T) (36.0) & 44.2 \\
SURYS & 2 & eDifFIQA(L) (31.4) & 44.0 \\
Twente & 4 & MagFace (34.3) & 45.0 \\
Improved Twente & 5 & MagFace (33.7) & 45.0 \\
LADIMO-inspired & 4 & ViT-FIQA(T) (20.9) & 39.7 \\
StyleDemorpher-inspired & 6 & eDifFIQA(L) (24.1) & 43.2 \\
Arc2Morph & 10 & ViTNT-FIQA (10.0) & 56.2 \\
StableMorph & 10 & ViTNT-FIQA (20.9) & 51.5 \\
\bottomrule
\end{tabular}
\end{table}

\FloatBarrier
\section{Score Distributions}
\label{app:distributions}
\begin{figure}[!htbp]
\centering
\includegraphics[width=\textwidth]{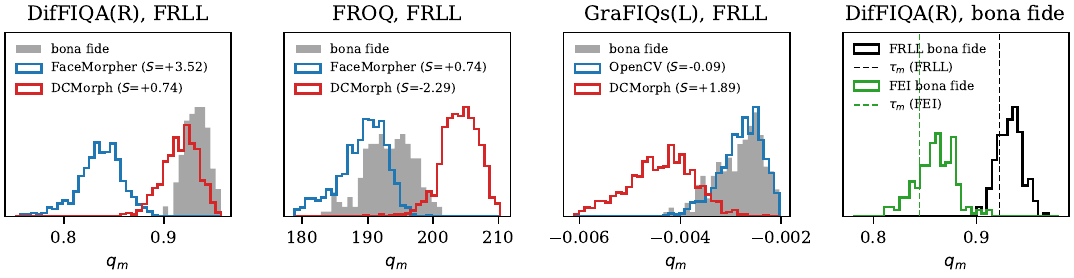}
\caption{Histograms of the direction-normalized scores $q_m$. From left to right: DifFIQA(R), FROQ, and GraFIQs(L) on the FRLL bona fide images and two attacks each, with the signed separation $S$ of Eq.~(\ref{eq:fdr}), and DifFIQA(R) on the bona fide images of FRLL and FEI with the threshold $\tau_m$ set at BSCER $=20\%$ on each dataset (dashed).}
\label{fig:app_distributions}
\end{figure}

Fig.~\ref{fig:app_distributions} shows the score distributions behind some results of Sec.~\ref{sec:results}. DifFIQA(R) gives FaceMorpher morphs much lower scores than bona fide images, and DCMorph morphs only slightly lower scores. FROQ gives DCMorph morphs higher scores than bona fide images, the largest reversal of the study. GraFIQs(L) separates DCMorph in the expected direction but not OpenCV. The bona fide scores of DifFIQA(R) are lower on FEI than on FRLL, so a threshold set on one dataset does not fit the other (Sec.~\ref{sec:transfer}).

\begin{figure}[!htbp]
\centering
\includegraphics[width=\textwidth]{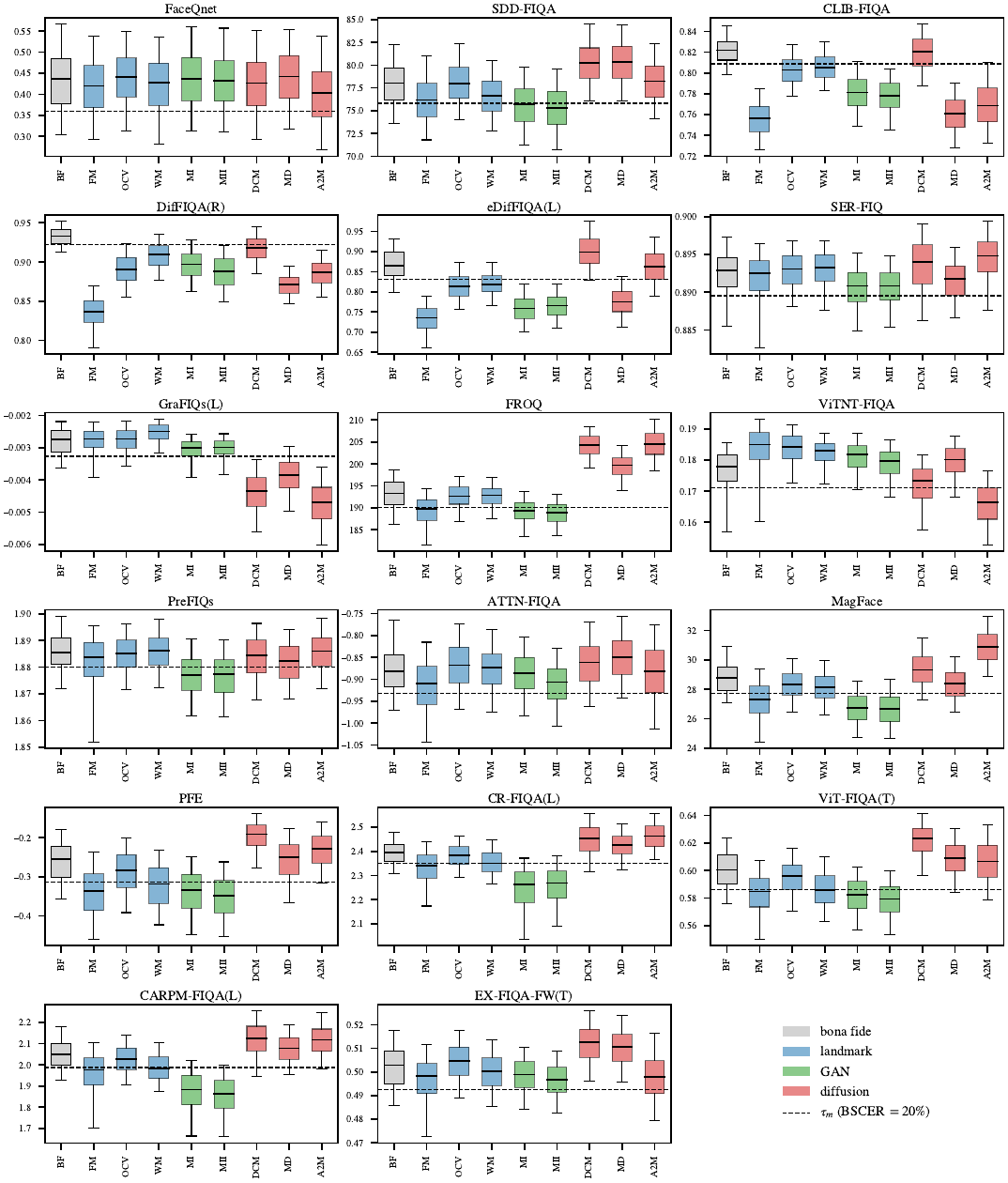}
\caption{Distribution of the direction-normalized score $q_m$ of every method on the FRLL bona fide images (BF) and each FRLL attack. The box spans the interquartile range, the line marks the median, and the whiskers mark the 5th and 95th percentiles. Dashed: threshold $\tau_m$ at BSCER $=20\%$. Attack abbreviations as in Tab.~\ref{tab:mad_results}.}
\label{fig:app_dist_frll}
\end{figure}

\begin{figure}[!htbp]
\centering
\includegraphics[width=\textwidth]{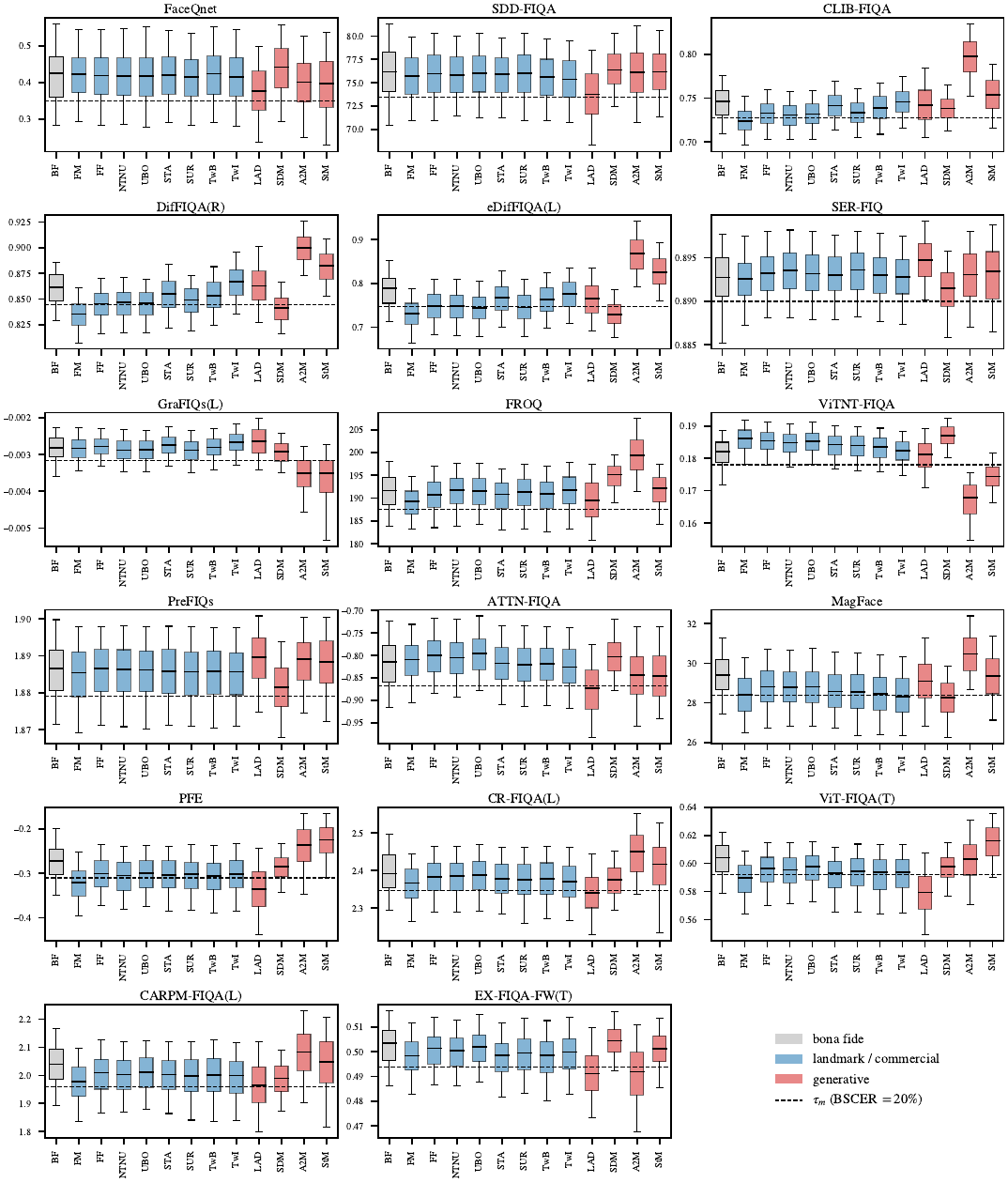}
\caption{Distribution of $q_m$ of every method on the FEI bona fide images (BF) and each FEI attack, as in Fig.~\ref{fig:app_dist_frll}. Attack abbreviations as in Tab.~\ref{tab:fei}.}
\label{fig:app_dist_fei}
\end{figure}

Figs.~\ref{fig:app_dist_frll} and~\ref{fig:app_dist_fei} show the score distributions of all methods on all attacks, with the threshold $\tau_m$ at BSCER $=20\%$. Attack scores below the bona fide scores mean that morphs receive lower quality, and attack scores above them are reversals. The score scales differ between methods.

\FloatBarrier

\end{document}